%% file: main.tex
\documentclass[11pt]{article}
\usepackage[preprint]{acl}
\usepackage{times}
\usepackage{latexsym}
\usepackage[T1]{fontenc}
\usepackage[utf8]{inputenc}
\usepackage{microtype}
\usepackage{inconsolata}
\usepackage{graphicx}

\input{math_commands.tex}

\usepackage{url}
\usepackage{booktabs}
\usepackage{amssymb}
\usepackage{array}
\usepackage{tabularx}
\usepackage{longtable}
\usepackage{multirow}
\usepackage{xcolor}
\usepackage{colortbl}

\title{Beyond the Transcript: Measuring and Improving the Use of Acoustic Information in Audio Language Models}

\author{
  Linkai Peng$^{1}$ \qquad Baorian Nuchged$^{2}$ \qquad Yuyang Yao$^{3}$ \\
  $^{1}$University of Connecticut \qquad
  $^{2}$The University of Texas at Austin \qquad
  $^{3}$University of Oxford \\
}

\begin{document}

\maketitle

\input{sec/00_abstract}
\input{sec/01_intro}
\input{sec/02_design}
\input{sec/03_gap}
\input{sec/04_diag}
\input{sec/05_repair}
\input{sec/06_related}
\input{sec/07_conclusion}
\input{sec/08_limitations}

\bibliography{refs}

\appendix
\input{sec/app_a_inputs}
\input{sec/app_b_classes}
\input{sec/app_c_systems}

\input{sec/app_e_readout_method}
\input{sec/app_f_readout_results}
\input{sec/app_g_training}

\input{sec/app_h_repair}

\end{document}

%% file: math_commands.tex
\usepackage{amsmath,amsfonts,bm}

\def\eqref#1{equation~\ref{#1}}
\def\1{\bm{1}}

\DeclareMathAlphabet{\mathsfit}{\encodingdefault}{\sfdefault}{m}{sl}
\SetMathAlphabet{\mathsfit}{bold}{\encodingdefault}{\sfdefault}{bx}{n}

%% file: sec/00_abstract.tex
\begin{abstract}
% 中文：音频语言模型基准通过多种任务提供全面评价。然而，部分任务仅凭转写甚至题目本身就能作答，因此基准分数可能混杂声学理解与语言建模能力。我们把转写通常足以作答的词汇类任务，与全局属性类和录音内部结构类任务区分开，后两类需要模型分析转写之外的录音本身，或从整体判断说话人与情绪等属性，或从内部判断重音与语调等模式。我们在六个基准的 92 个任务上评测了 19 个开源模型和 Gemini 3.1 Pro，并通过把原录音与其转写的合成朗读比较、以及把原录音加到转写之上，估计录音增益，即原录音相对保留其词的参照的准确率优势。在属性类和结构类任务上，开源模型落后 Gemini 28.6 和 41.9 个百分点，每类差距的一半以上落在录音增益上。尽管如此，Gemma-4 12B 和 Qwen2.5-Omni-7B 中的线性探针表明，作答所需的声学信息在语言模型早期层已可读出，而多种线索的读出在随后的层中下降。因此，我们在这些层中训练 LoRA 适配器并加入内容锚，两个模型的属性类和结构类准确率提高 12.8 至 44.4 个百分点，平均词汇准确率与基础模型相差不到 1 个百分点。大部分提升体现为录音增益的增大。因此，训练改善的是模型对其已编码声学信息的利用。
Audio language model benchmarks provide comprehensive evaluation across diverse tasks. However, some tasks can be answered from the transcript or even the question alone, and benchmark scores can therefore confound acoustic understanding with language modeling ability. We distinguish lexical tasks, for which the transcript usually suffices, from global-attribute and within-recording-structure tasks, which require the model to analyse the recording beyond its words, either as a whole, for properties such as the speaker and emotion, or within it, for patterns such as stress and intonation. We evaluated 19 open models and Gemini 3.1 Pro on 92 tasks from 6 benchmarks and estimated the recording gain, the accuracy advantage of the original recording over references that retain its words, by comparing the recording with synthetic readings of its transcript and by adding it to the transcript. On attribute and structure tasks, the open models fall behind Gemini by 28.6 and 41.9 percentage points, and about half or more of each deficit lies in the recording gain. Nevertheless, linear probes in Gemma-4 12B and Qwen2.5-Omni-7B show that the acoustic information needed for answering is already readable in early language-model layers, while the readout of many cues declines in the layers that follow. We therefore constructed training data and trained LoRA adapters in these layers, with a content anchor, which raises attribute and structure accuracy by 12.8 to 44.4 points in the two models and keeps mean lexical accuracy within 1 point of the base model. Most of the improvement appears as larger recording gains. These results suggest that training improves the use of acoustic information that the models already encode.
\end{abstract}

%% file: sec/01_intro.tex
% 中文：引言
\section{Introduction}
\label{sec:intro}

% 中文：音频语言模型将大语言模型的语言理解与推理能力扩展到语音及其他音频输入 \citep{xu2025qwen25omni}。综合基准通过从口语内容理解到说话人、情绪和韵律识别等任务评估这些能力 \citep{wang2024audiobench,yang2024airbench,sakshi2024mmau,wang2026mmsu}。然而，即使输入是语音，模型也可能仅凭词汇内容，甚至题目与选项本身，回答部分问题。例如，仅凭「我太高兴了」的措辞就可能足以回答情绪题，无需分析音频中的语气。在这种情况下，级联系统（自动语音识别后接文本语言模型）可能与音频语言模型表现相当。因此，基准分数可能混杂语言建模能力与声学理解。那么，音频语言模型从原录音中获得的帮助，在这些已识别的词之外还有多少？
Audio language models extend the language understanding and reasoning capabilities of large language models to speech and other audio inputs \citep{xu2025qwen25omni}. Comprehensive benchmarks assess these capabilities through tasks ranging from spoken-content understanding to speaker, emotion and prosody recognition \citep{wang2024audiobench,yang2024airbench,sakshi2024mmau,wang2026mmsu}. However, even with speech as input, a model may rely only on its lexical content, or even the question and options alone, to answer some questions. For example, the wording of \emph{I am so happy} may suffice to answer an emotion question without analysing the tone of the audio. In this situation, a cascaded system (automatic speech recognition followed by a text-based language model) may perform as well as an audio language model. Benchmark scores can therefore confound language modeling ability with acoustic understanding. How much additional benefit does an audio language model obtain from the original recording beyond these recognized words?
% 中文：对于仅凭词无法回答的问题，模型必须从录音本身提取所需的线索，而这些线索因题目而异。因此，我们依据作答所需的证据区分三类任务（图~\ref{fig:method} 上方）。词汇类任务通常仅凭 ASR 转写即可作答。全局属性类任务考察整段录音的属性，例如说话人身份与情绪，而录音内部结构类任务考察韵律模式和片段间的关系，例如重音与语调。这些结构类问题需要定位录音中的相关部分，并判断它们是如何说出的，例如哪个词被强调，或两段中哪段音高更高。由于 ASR 转写通常没有明确记录这些声学模式，原录音在属性类和结构类任务上能多给的最多，在词汇类上最少。 

\begin{figure*}[t]
\centering
\includegraphics[width=\linewidth]{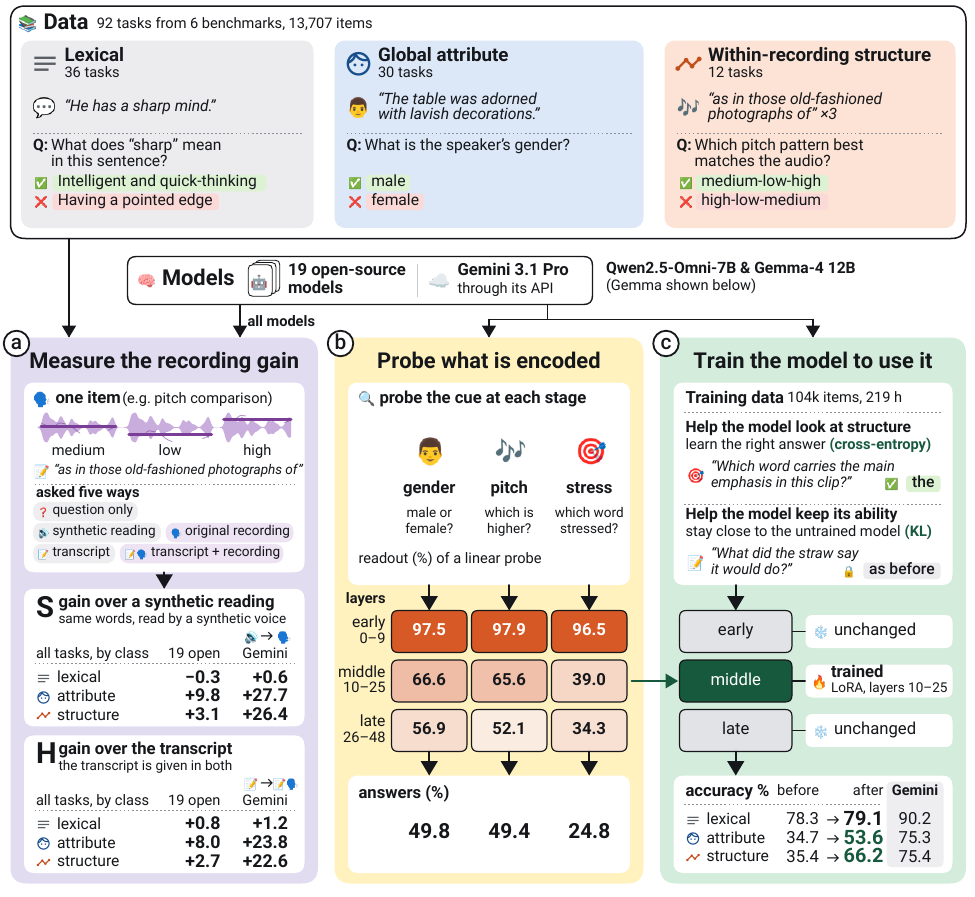}
% 中文：研究总览，从上方的数据和模型，到录音增益（a）、读出（b）与修复（c）。任务按答案所需证据的位置分组。词汇类可从词中作答，全局属性类依赖说话人等整段录音的属性，录音内部结构类依赖录音内的关系，例如音高模式。每类展示一道示例题，14 个混合类任务未画出。（a）一道题的五种问法，以及 19 个开源模型的平均值与 Gemini 3.1 Pro 在各类上的 S 与 H（百分点）。（b）Gemma-4 12B 早期、中间与晚期层对三种声学线索的读出，以及同一批题目上的作答准确率（随机水平为 50%、50% 和 25%）。每个单元格报告所示层段内某一层的读出。（c）在读出下降的层上进行 LoRA 训练（声学题用交叉熵，内容题用 KL 散度贴近未训练的模型），以及训练前后的各类准确率，Gemini 的结果作为参照。
\caption{Overview of the study, from the data and models (top) to the recording gain (a), the readout (b) and the repair (c). The tasks are grouped by where the evidence for the answer lies. Each class is shown with one example item, and the 14 mixed tasks are not shown. (a) One item asked with five inputs, and the class means of $S$ and $H$ in percentage points (pp) for the mean of the 19 open models and for Gemini 3.1 Pro. (b) Readout of three acoustic cues in the early, middle and late layers of Gemma-4 12B, and answer accuracy on the same items (chance 50\%, 50\% and 25\%). (c) LoRA training of the layers where readout declines, with cross-entropy on acoustic questions and a KL term, and accuracy by class before and after training, with Gemini shown for reference.}
\label{fig:method}
\end{figure*}

The answer is likely to differ across tasks.  For questions that the words alone cannot answer, the model must extract the required cues from the recording itself, and these cues vary with the question. We therefore distinguish three task classes by the evidence needed to answer them (top of Figure~\ref{fig:method}). \emph{Lexical} tasks can usually be answered using the ASR transcript alone. \emph{Global-attribute} tasks assess properties of the recording as a whole, such as speaker identity and emotion, while \emph{within-recording structure} tasks assess prosodic patterns and relations between segments, such as stress and intonation. These questions require locating relevant parts of the recording and reasoning how they were spoken, for example which word was emphasized or which of two segments had a higher pitch. Because an ASR transcript usually leaves these acoustic patterns unspecified, the recording has most to add on attribute and structure tasks and least on lexical tasks.

% 中文：要检验模型是否用上了这些证据，需要一个保留词汇内容的参照。加噪、静音和去除音频会同时破坏词汇内容与其他声学线索 \citep{sakshi2024mmau,wang2026mmsu,he2026measuring,foo2026glitters}。让词与声音冲突的测试能揭示模型跟随哪一方，但需要另行构造刺激 \citep{chen2025listen,xiong2026deaf,pang2026voxparadox}。我们在既有基准题目上通过两种比较估计原录音在转写之外的贡献（图~\ref{fig:method}a，第~\ref{sec:design} 节），将相应的准确率差称为录音增益。S 是原录音相对其转写的合成朗读的准确率优势，朗读保留转写中的词，替换声音和韵律，并移除背景声。H 是在转写之外加入原录音带来的准确率增益。两种比较涉及不同的输入变化，我们考察两者共同呈现的规律。我们在六个基准的 92 个任务、13,707 道题上评测了 19 个开源模型和作为闭源参照的 Gemini 3.1 Pro，随后在两个开源模型中比较线性探针准确率（简称读出）与作答准确率，并检验一种训练修复。这些测量共同表明，开源模型编码的声学信息多于它们用到的。下面三条发现依次给出证据（图~\ref{fig:method}）。
Measuring whether models use this evidence requires a reference that retains the words. Noise, silence and audio removal disrupt the words along with other acoustic cues \citep{sakshi2024mmau,wang2026mmsu,he2026measuring,foo2026glitters}, and tests that place the words in conflict with the voice require newly constructed stimuli \citep{chen2025listen,xiong2026deaf,pang2026voxparadox}. We estimate the contribution of the original recording beyond the transcript using two comparisons on existing benchmark items (Figure~\ref{fig:method}a and Section~\ref{sec:design}). We call the corresponding accuracy differences the \emph{recording gains}. $S$ is the recording's accuracy advantage over synthetic readings of its transcript, which retain the transcribed words, replace the voice and prosody and remove background sound. $H$ is the accuracy gained by adding the recording to its transcript. The comparisons involve different changes to the input, and we examine the patterns they share. We evaluate 19 open models and, as a closed-model reference, Gemini 3.1 Pro on 13,707 items from 92 tasks in 6 benchmarks, then compare linear-probe \emph{readout}, with answer accuracy in two open models and test a repair. In these two models examined internally, the measurements show that acoustic information can be present in the representations but underused in answers. Three findings give the evidence (Figure~\ref{fig:method}).

\begin{itemize}
% 中文：开源模型落后 Gemini 最多的是属性类和结构类任务，平均 28.6 和 41.9 个百分点，这一差距的一半以上落在录音增益上（第~\ref{sec:gap} 节）。因此，开源模型从录音信息中获得的帮助少于 Gemini，后者的 S/H 比开源模型高 17.9/15.8 和 23.3/19.9 个百分点。
\item \textbf{Recording gain.} The open models fall behind Gemini most on attribute and structure tasks, by 28.6 and 41.9\,pp on average, and about half or more of this deficit lies in the recording gain (Section~\ref{sec:gap}). The open models thus benefit less from the information in the recording than Gemini, whose $S$/$H$ exceed theirs by 17.9/15.8 and 23.3/19.9\,pp.

% 中文：在无独立编码器的 Gemma-4 12B \citep{gemmateam2026gemma4} 和带编码器的 Qwen2.5-Omni-7B \citep{xu2025qwen25omni} 中，多种属性类和结构类任务所需的声学信息都能从早期层读出，作答准确率却远低于读出（第~\ref{sec:diag} 节）。读出与作答之间的这种差距称为利用缺口。
\item \textbf{Readout.} In the encoder-free Gemma-4 12B \citep{gemmateam2026gemma4} and the encoder-based Qwen2.5-Omni-7B \citep{xu2025qwen25omni}, the acoustic information needed for many attribute and structure tasks is readable from early layers, while answer accuracy stays well below this readout (Section~\ref{sec:diag}). We call this difference the \emph{utilization gap}.

% 中文：我们构建新的训练数据，在读出下降的层上训练低秩适配器（LoRA，\citealp{hu2022lora}），并用内容锚使内容题的回答贴近基础模型（第~\ref{sec:repair} 节）。训练使两个模型的属性类和结构类准确率提高 12.8 至 44.4 个百分点，平均词汇准确率与基础模型相差不到 1 个百分点。在结构类任务上，两个模型的 S/H 从 2.5/5.2 和 1.8/2.7 升至 29.6/29.5 和 38.9/46.0 个百分点。训练后的读出与适配器分析表明，这一提升来自对表示中已经编码的声学信息的更好利用。
\item \textbf{Repair.} We construct new training data and train low-rank adapters (LoRA, \citealp{hu2022lora}) in the layers where readout declines, with a content anchor that keeps answers to content questions close to the base model's (Section~\ref{sec:repair}). Training improves attribute and structure accuracy by 12.8 to 44.4\,pp in the two models, while mean lexical accuracy remains within 1\,pp of the base model. On structure tasks, $S$/$H$ rise from 2.5/5.2 and 1.8/2.7 to 29.6/29.5 and 38.9/46.0\,pp. Post-training analyses of readout and of the adapters are consistent with better use of acoustic information already encoded in the representations.
\end{itemize}

% 中文：我们公开逐题作答结果、生成合成朗读的脚本，以及重新计算本文所有数字和图表的代码。
We release the data-construction scripts, the code and the trained LoRA adapters.

%% file: sec/02_design.tex
% 中文：测量录音增益
\section{Measuring the Recording Gain}
\label{sec:design}

% 中文：我们依据任务说明和答案形式，按作答所需的证据对基准任务分类。评测包括来自六个基准的 92 个任务、13,707 道题（AudioBench \citep{wang2024audiobench}、AIR-Bench \citep{yang2024airbench}、Dynamic-SUPERB Phase-2 \citep{huang2025dynamicsuperb}、MMAU \citep{sakshi2024mmau}、MMAU-Pro \citep{kumar2025mmaupro} 和 MMSU \citep{wang2026mmsu}，附录~\ref{app:a}）。36 个词汇类任务通常凭转写即可作答。30 个全局属性类任务需要整段录音的证据，例如说话人、情绪、场景或说话人数；12 个录音内部结构类任务需要关于事件位置或片段间关系的证据，例如哪个词有重音、两段中哪段音高更高。其余 14 个混合类任务结合词汇与声学判断，或需要额外推理。第~\ref{sec:gap} 至~\ref{sec:repair} 节的分类结果覆盖前三类的 78 个任务、12,102 道题，混合类任务只出现在附录~\ref{app:h} 的逐任务结果中。各类包含的任务见附录~\ref{app:b}。
We classify benchmark tasks by the evidence needed to answer them, based on each task's description and answer format. Our evaluation covers 92 tasks and 13,707 items from AudioBench \citep{wang2024audiobench}, AIR-Bench \citep{yang2024airbench}, Dynamic-SUPERB Phase-2 \citep{huang2025dynamicsuperb}, MMAU \citep{sakshi2024mmau}, MMAU-Pro \citep{kumar2025mmaupro} and MMSU \citep{wang2026mmsu} (Appendix~\ref{app:a}). The transcript usually suffices for the 36 \textbf{lexical} tasks. The 30 \textbf{global-attribute} tasks need evidence about the recording as a whole, such as the speaker, emotion, scene or number of speakers, and the 12 \textbf{within-recording structure} tasks need evidence about where something occurs or how segments relate, such as which word is stressed or which of two segments is higher in pitch. The remaining 14 \textbf{mixed} tasks combine lexical and acoustic judgements or require additional reasoning. The class results in Sections~\ref{sec:gap} to~\ref{sec:repair} cover the 78 tasks and 12,102 items of the first three classes, and the mixed tasks appear only in the per-task results of Appendix~\ref{app:h}. Appendix~\ref{app:b} lists the tasks in each class.
\begin{table}[h]
% 中文：五种输入条件。
\caption{The five input conditions.}
\label{tab:inputs}
\vspace{4pt}
\centering
\small
\renewcommand{\arraystretch}{1.15}
\setlength{\tabcolsep}{4pt}
\newcommand{\emo}[1]{\raisebox{-0.28em}{\includegraphics[height=1.15em]{figs/emoji/noto_#1.png}}}
\begin{tabular}{@{}c@{\hspace{4pt}}lccc@{}}
\toprule
 & & & \multicolumn{2}{c@{}}{Form} \\
\cmidrule(l){4-5}
\multicolumn{2}{@{}l}{Input} & Acc. & Text & Speech \\
\midrule
 & Question and options only & $N$ & & \\
\emo{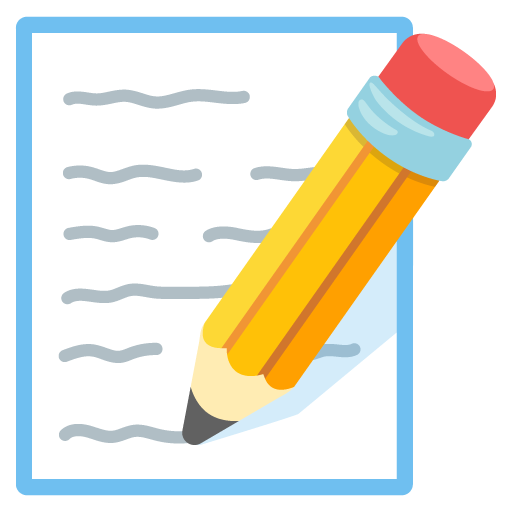} & Whisper transcript & $A$ & $\checkmark$ & \\
\emo{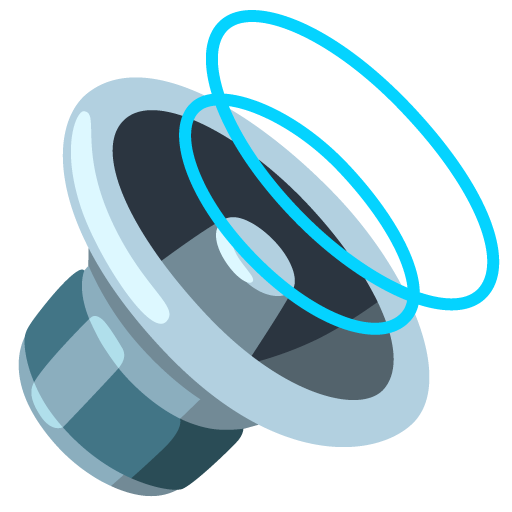} & Synthetic reading of it & $T$ & & $\checkmark$ \\
\emo{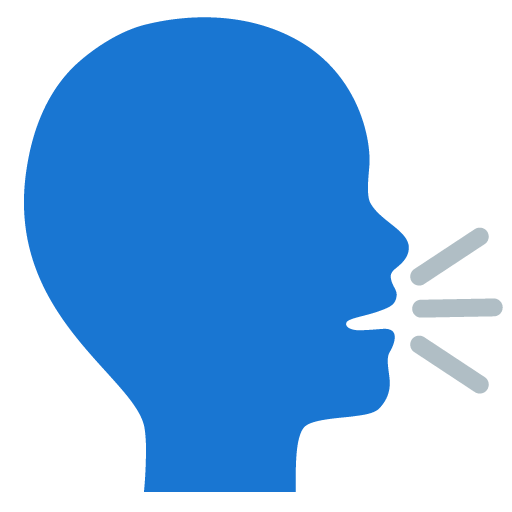} & Original recording & $O$ & & $\checkmark$ \\
\emo{1f4dd}\emo{1f5e3} & Transcript and recording & $AO$ & $\checkmark$ & $\checkmark$ \\
\bottomrule
\end{tabular}
\end{table}

% 中文：为了估计原录音在转写之外的贡献，我们为每道题构造五种输入（表~\ref{tab:inputs}），题目和选项保持不变。$N$ 是只给题目和选项时的准确率。$A$ 是给 Whisper \citep{radford2022whisper} 对原录音的转写时的准确率。$T$ 是给转写的合成朗读时的准确率，$O$ 是给原录音时的准确率，$AO$ 是同时给转写与原录音时的准确率。两种录音增益分别为 $S=O-T$，即原录音相对同一转写的合成朗读的准确率优势，以及 $H=AO-A$，即在转写之外加入原录音带来的准确率增益。语音合成系统 \citep{edgetts} 用一个女声和一个男声分别朗读转写，$T$ 是两种朗读上的平均准确率。朗读保留转写中的词，替换原录音的声音和韵律，并移除背景声。例如，若模型只根据合成声音回答二选一的性别题，两种朗读上的平均准确率为随机水平。保留下来的朗读相对源转写的词错率为 1.4\%。
To estimate the contribution of the original recording beyond the transcript, we pose each item under five inputs with the same question and options (Table~\ref{tab:inputs}). $N$ is accuracy with only the question and options. $A$ is accuracy with the transcript of the recording produced by Whisper \citep{radford2022whisper}. $T$ is accuracy with a synthetic reading of the transcript, $O$ with the original recording and $AO$ with the transcript and recording together. The two recording gains are $S=O-T$, the recording's advantage over synthetic readings of the same transcript, and $H=AO-A$, the accuracy gained by adding the recording to its transcript. The text-to-speech system \citep{edgetts} reads the transcript once in a female and once in a male voice, and $T$ is the mean over these two readings. The readings retain the transcribed words while replacing the original voice and prosody and removing background sound. A model answering a binary gender question from the synthetic voice alone, for example, averages chance accuracy over the two readings. The retained readings have a word error rate of 1.4\% relative to their source transcripts.
\begin{figure*}[t]
\centering
\includegraphics[width=\linewidth]{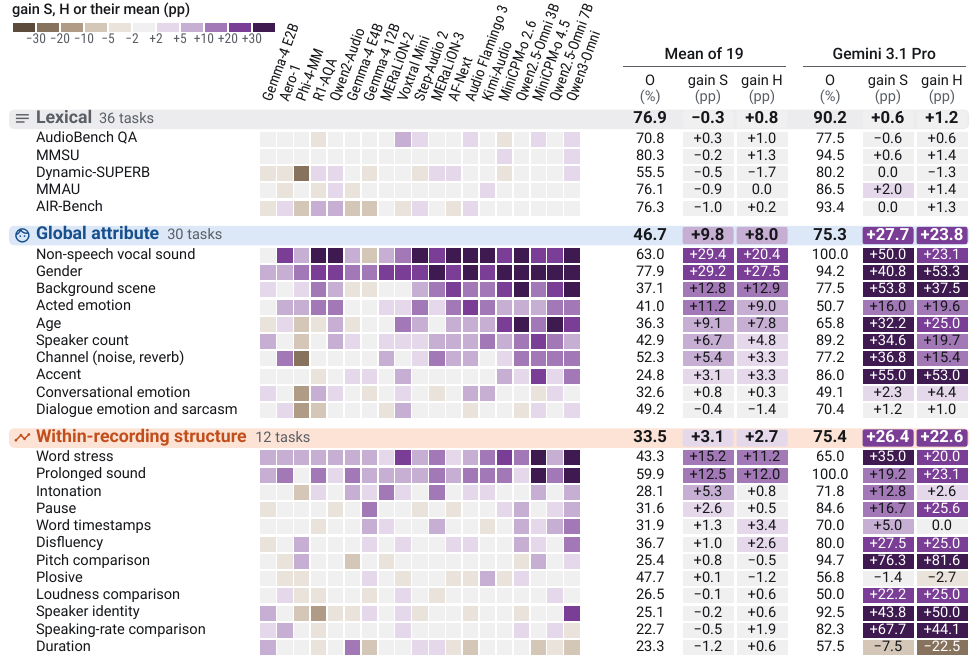}
% 中文：开源模型以及 Gemini 3.1 Pro 在三种任务类型上的录音增益。每行汇总一类中的一组任务（部分任务）。（左侧） 19 列为开源模型，按原录音上的总体准确率从低到高排列，单元格颜色表示该模型的 S 与 H 均值。（右侧） 19 个开源模型均值与 Gemini 3.1 Pro 在原录音上的准确率 O（%）和两种增益（百分点） S，以及 H。紫色表示正增益，棕色表示负增益。混合类任务未画出。
\caption{Recording gains of the open models and Gemini 3.1 Pro on the three task classes. Each row summarizes a group of tasks within a class. On the left, the 19 columns represent the open models, ordered from lowest to highest overall accuracy with the original recording, and each cell shows that model's mean of $S$ and $H$. The right side reports accuracy with the original recording, $O$ (\%), and the two gains (pp), $S$ and $H$, for the mean of the 19 open models and for Gemini 3.1 Pro. Purple marks positive gains and brown negative gains. Mixed tasks are not shown.}
\label{fig:smatrix}
\end{figure*}
% 中文：理想情况下，每种比较的两个输入只在词之外的声音上有差别。但每种比较各多出一处差别。S 以合成语音为参照，因此也反映自然语音与合成语音之间的差别。H 的两个条件都给了转写，因此也反映模型如何把文字与音频结合。这两处差别各自只进入一个量，单独任何一处都造不出 S 和 H 都显示的差异。这样的共同差异来自词之外的声音，我们的结论只建立在这样的差异上。两种比较依赖同一份转写，其错误与遗漏可能影响估计（见“局限”一节）。我们在全部题目上评测了 19 个开源端到端音频语言模型，并以 Gemini 3.1 Pro 作为闭源参照（附录~\ref{app:c}）。准确率先在任务内按题平均，再在类别内对任务等权平均，开源模型均值对 19 个模型等权。
Ideally, the two inputs of each comparison differ only in the sound beyond the words. Each comparison, however, adds one further difference. $S$ uses synthetic speech as its reference and therefore also reflects differences between natural and synthetic speech. $H$ gives the model the transcript in both conditions and therefore also reflects how the model combines text with audio. Neither of these differences enters the other measure, and neither alone can create a difference that both $S$ and $H$ show. Such shared differences measure the recording gain relative to transcript-based references, and our conclusions rest on them (see Limitations). We evaluated 19 open end-to-end audio language models on all items, with Gemini 3.1 Pro as a closed-model reference (Appendix~\ref{app:c}). Accuracy is averaged over items within each task and then equally over tasks within each class, and open-model means weight the 19 models equally. 

%% file: sec/03_gap.tex
% 中文：按模型与任务类别的录音增益
\section{Recording Gain by Model and Class}
\label{sec:gap}

% 中文：图~\ref{fig:smatrix} 报告 19 个开源模型和 Gemini 的原录音准确率以及 $S$、$H$。开源模型在三类任务中均落后，平均准确率在词汇类、属性类和结构类分别低 13.3、28.6 和 41.9 个百分点。即使是最强的开源模型 Qwen3-Omni，在属性类和结构类上也只有 59.9% 和 49.8%。

% 在词汇任务上，两种比较下的增益都很小，开源模型均值与 Gemini 的平均 S、H 均落在 −0.3 至 +1.2 个百分点之间。这一结果符合预期，词汇任务不太依赖录音中词之外的信息。Gemini 的词汇类准确率也更高，反映的是其语言模型的优势。
Figure~\ref{fig:smatrix} reports accuracy with the original recording and the two gains for the 19 open models and Gemini. The open models trail Gemini in all three classes, by 13.3, 28.6 and 41.9\,pp on lexical, attribute and structure tasks on average, and even the strongest open model, Qwen3-Omni, reaches only 59.9\% and 49.8\% on attribute and structure tasks. 

On lexical tasks, gains under both comparisons are small. Mean $S$ and $H$ range from $-0.3$ to $+1.2$\,pp for the open-model mean and Gemini. This pattern is expected, because lexical tasks depend little on information in the recording beyond the words. Gemini's higher lexical accuracy therefore reflects the strength of its language model.

% 中文：在属性类和结构类任务上，这一差距的一半以上来自录音增益之差。给原录音时，Gemini 的准确率为 75.3% 和 75.4%，开源模型均值为 46.7% 和 33.5%，差距 28.6 和 41.9 个百分点。S 之差为 27.7 − 9.8 = 17.9 和 26.4 − 3.1 = 23.3 个百分点，H 之差为 23.8 − 8.0 = 15.8 和 22.6 − 2.7 = 19.9 个百分点，都超过差距的一半（附录~\ref{app:d}）。最强的开源模型 Qwen3-Omni 的 S/H 为 18.4/14.7 和 12.0/10.1 个百分点，也低于 Gemini。这些较小的录音增益促使我们考察，作答所需的声学信息是模型内部没有，还是有而没有用上。
On attribute and structure tasks, about half or more of this deficit is a difference in recording gain. With the original recording, Gemini reaches 75.3\% and 75.4\%, against 46.7\% and 33.5\% for the open-model mean, gaps of 28.6 and 41.9\,pp. $S$ differs by $27.7-9.8=17.9$ and $26.4-3.1=23.3$\,pp, and $H$ by $23.8-8.0=15.8$ and $22.6-2.7=19.9$\,pp, in each case about half or more of the gap. Qwen3-Omni, the strongest open model, has $S$/$H$ of 18.4/14.7 and 12.0/10.1\,pp, also below Gemini. These small recording gains lead us to ask whether the acoustic information needed for the answers is absent from the models or present but unused.

%% file: sec/04_diag.tex
% 中文：逐层读出
\section{Layer-wise Readout}
\label{sec:diag}

\begin{figure*}[t]
\centering
\includegraphics[width=\linewidth]{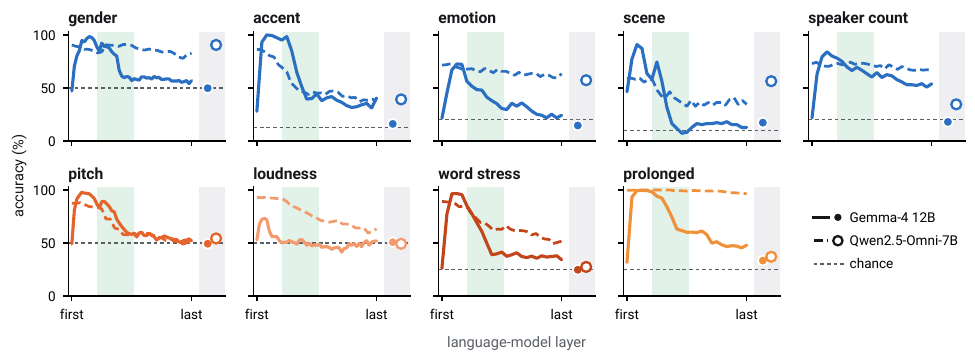}
% 中文：Gemma-4 12B 与 Qwen2.5-Omni-7B 的逐层线性探针读出与作答准确率。圆点表示原录音上的作答准确率。绿色带标出 Gemma-4 12B 的读出下降层（第 10–25 层）。蓝色面板为五种属性任务，橙色面板为四种结构任务。
\caption{Layer-wise linear-probe readout and answer accuracy in Gemma-4 12B and Qwen2.5-Omni-7B. The dots show answer accuracy with the original recording. The green band marks the readout-decline layers of Gemma-4 12B (layers 10--25). Blue panels show five attribute tasks and orange panels four structure tasks.}
\label{fig:probe}
\end{figure*}

% 中文：为了度量模型的表示中是否含有作答所需的声学信息，我们在冻结模型每一层的表示上训练线性探针，并把探针的准确率称为读出。属性类任务上，探针在评测题上训练和测试，以任务标签为目标。结构类任务上，探针需要知道被比较的部分在录音中的位置，因此我们构造了录音，例如把同一说话人的两段话拼接并让两段的音高、语速或响度不同。探针得到各部分的边界，对每部分内的表示取平均，再由两者之差判断哪部分的值更大（附录~\ref{app:e}）。此外我们还让模型直接作答得到回答准确率。若探针能从表示中读出答案而模型答错，说明这些信息在表示中存在，却没有进入模型的作答。
To measure whether a model's representations contain the acoustic information needed for answering, we train a linear probe on the representations of each layer of the model and call the probe's accuracy the \emph{readout}. For attribute tasks, the probe is trained and tested on the evaluation items with the task labels as targets. For structure tasks, the probe needs to know where the compared parts lie, and we therefore constructed recordings, for example by joining two utterances of the same speaker that differ in pitch, speaking rate or loudness. The probe receives the segment boundaries, averages the representations within each segment and uses their difference to judge which has the larger value (Appendix~\ref{app:e}). The models also answer these items directly, which gives answer accuracy. If a probe reads the answer from the representations while the model answers incorrectly, the information is present but not reliably used in the answer.

% 中文：我们检查 Qwen2.5-Omni-7B 与 Gemma-4 12B。前者拥有独立音频编码器，后者将波形帧直接投影到语言模型中，这使我们能分别检查有、无独立编码器时的利用缺口。此外，Qwen2.5-Omni-7B 属于表现最好的开源模型之一，在属性类和结构类任务上的原录音准确率为 56.7% 和 34.0%，高于开源模型均值 46.7% 和 33.5%，按总体准确率在 19 个开源模型中排第二（图~\ref{fig:smatrix}）。Gemma-4 12B 排第 13，为 34.7% 和 35.4%。
We examine Qwen2.5-Omni-7B, which has a separate audio encoder, and Gemma-4 12B, which projects waveform frames directly into the language model, to study the utilization gap with and without an encoder. In addition, Qwen2.5-Omni-7B is among the strongest open models, reaching 56.7\% and 34.0\% with the original recording on attribute and structure tasks, above the open-model means of 46.7\% and 33.5\%, and ranking second of the 19 by overall accuracy (Figure~\ref{fig:smatrix}). Gemma-4 12B ranks 13th, reaching 34.7\% and 35.4\%.

% 中文：图~\ref{fig:probe} 比较两个模型的逐层读出和作答准确率（全部探针任务见附录~\ref{app:f}）。在两个模型的所有探针任务上，读出的峰值都高于作答准确率，并且多数任务的峰值出现在早期层。以音高为例，Gemma-4 12B 第 4 层的读出达到 97.9%，Qwen2.5-Omni-7B 第 2 层达到 88.0%，而作答准确率只有 49.4% 和 54.3%（随机为 50%）。后层读出下降或保持较高时，都可能伴随较差的作答。在 Gemma-4 12B 中，性别、口音、情绪、场景、音高以及找出被重读词的读出都在第 10–25 层下降。在 Qwen2.5-Omni-7B 中，口音、场景、音高、响度以及找出被重读词的读出都在第 2–11 层下降。相比之下，找出被拉长词的读出到最后一层仍为 96.2%，作答准确率只有 37.1%（随机为 25%）。我们依据多个读出下降的范围，选择 Gemma-4 12B 的第 10–25 层和 Qwen2.5-Omni-7B 的第 2–11 层，称为读出下降层。（脚注：图~\ref{fig:method} 把 Gemma-4 12B 的层分为早期、中期和晚期，其中的中期层就是该模型的读出下降层。）第~\ref{sec:repair} 节检验在这里训练能否在保持平均词汇准确率的同时改善作答，以及提升是否来自对已可读信息的利用。
Figure~\ref{fig:probe} compares layer-wise readout with answer accuracy (all probed tasks in Appendix~\ref{app:f}). In both models, peak readout exceeds answer accuracy on every probed task, and for most tasks the peak lies in the early layers. For pitch, for example, readout reaches 97.9\% at layer 4 of Gemma-4 12B and 88.0\% at layer 2 of Qwen2.5-Omni-7B, while answer accuracy is 49.4\% and 54.3\% (chance 50\%). Poor answers occur both when later-layer readout declines and when it stays high. In Gemma-4 12B, readouts of gender, accent, emotion, scene, pitch and the stressed word all fall across layers 10--25. In Qwen2.5-Omni-7B, readouts of accent, scene, pitch, loudness and the stressed word fall across layers 2--11. Readout of which word is prolonged, by contrast, is still 96.2\% at the last layer, while answer accuracy is 37.1\% (chance 25\%). From the ranges over which several readouts fall, we chose layers 10--25 in Gemma-4 12B and 2--11 in Qwen2.5-Omni-7B, the \emph{readout-decline layers}.\footnote{Figure~\ref{fig:method} labels the layers of Gemma-4 12B as early, middle and late, and its middle layers are the readout-decline layers of this model.} Section~\ref{sec:repair} tests whether training there improves answers while preserving mean lexical accuracy, and whether the improvement comes from using information that is already readable.

%% file: sec/05_repair.tex
% 中文：修复利用缺口
\section{Repairing the Utilization Gap}
\label{sec:repair}

\begin{table*}[t]
% 中文：Gemma-4 12B 与 Qwen2.5-Omni-7B 的基础模型和修复后模型在三类任务上的准确率与两种录音增益，以及 19 个开源模型均值与 Gemini 3.1 Pro。O 为原录音上的准确率（%），S、H 及变化量的单位为百分点。
\caption{Accuracy and the two recording gains by task class for the base and repaired Gemma-4 12B and Qwen2.5-Omni-7B, with the mean of the 19 open models and Gemini 3.1 Pro. $O$ is accuracy (\%) with the original recording, and $S$, $H$ and changes are in pp.}
\label{tab:repair-class}
\vspace{4pt}
\centering
\small
\setlength{\tabcolsep}{7.5pt}
\makeatletter % keep the row tint inside the rules: no overhang at the outer edges
\providecommand\tintL{\gdef\CT@cell@color{\@tempdimb\z@\global\let\CT@cell@color\relax}}
\providecommand\tintR{\gdef\CT@cell@color{\@tempdimc\z@\global\let\CT@cell@color\relax}}
\makeatother
\begin{tabular}{@{}l*{7}{>{\raggedleft\arraybackslash}p{3.0em}}@{}}
\toprule
% 中文：表中类别为词汇类、全局属性类和录音内部结构类。各模型依次报告基础模型、修复后模型与变化量。末尾列出 19 个开源模型均值和 Gemini 3.1 Pro 的结果。
 & \multicolumn{1}{@{}c@{}}{Lexical} & \multicolumn{3}{c}{Global attribute} & \multicolumn{3}{c@{}}{Within-recording structure} \\
\cmidrule(lr){2-2}\cmidrule(lr){3-5}\cmidrule(l){6-8}
 & \multicolumn{1}{c}{$O$} & \multicolumn{1}{c}{$O$} & \multicolumn{1}{c}{$S$} & \multicolumn{1}{c}{$H$} & \multicolumn{1}{c}{$O$} & \multicolumn{1}{c}{$S$} & \multicolumn{1}{c@{}}{$H$} \\
\midrule
\multicolumn{8}{@{}l}{\textbf{Gemma-4 12B}} \\
\quad Base & 78.3 & 34.7 & $-0.4$ & $+5.2$ & 35.4 & $+2.5$ & $+5.2$ \\
\quad Repaired & 79.1 & 53.6 & $+12.6$ & $+15.7$ & 66.2 & $+29.6$ & $+29.5$ \\
\rowcolor{black!8}
\tintL\quad Change & \textbf{+0.8} & \textbf{+18.9} & $+13.0$ & $+10.5$ & \textbf{+30.8} & $+27.0$ & \tintR$+24.3$ \\
\midrule
\multicolumn{8}{@{}l}{\textbf{Qwen2.5-Omni-7B}} \\
\quad Base & 83.4 & 56.7 & $+18.3$ & $+14.9$ & 34.0 & $+1.8$ & $+2.7$ \\
\quad Repaired & 83.7 & 69.5 & $+30.3$ & $+28.9$ & 78.4 & $+38.9$ & $+46.0$ \\
\rowcolor{black!8}
\tintL\quad Change & \textbf{+0.3} & \textbf{+12.8} & $+12.0$ & $+14.1$ & \textbf{+44.4} & $+37.2$ & \tintR$+43.3$ \\
\midrule
Mean of 19 open models & 76.9 & 46.7 & $+9.8$ & $+8.0$ & 33.5 & $+3.1$ & $+2.7$ \\
Gemini 3.1 Pro & 90.2 & 75.3 & $+27.7$ & $+23.8$ & 75.4 & $+26.4$ & $+22.6$ \\
\bottomrule
\end{tabular}
\end{table*}

% 中文：在读出下降层训练
\subsection{Training in the Readout-Decline Layers}
\label{sec:method}

% 中文：我们在读出下降层训练 LoRA 适配器，即 Gemma-4 12B 的第 10–25 层和 Qwen2.5-Omni-7B 的第 2–11 层。训练集包含 103,878 道监督题、共 219.0 小时音频，覆盖 34 个属性类和结构类评测任务的题型。录音来自 17 个语料库，约 80% 来自 LibriSpeech 的训练划分 \citep{panayotov2015librispeech}，其余来自 Expresso \citep{nguyen2023expresso}、Speech Accent Archive \citep{weinberger2015accent}、表演情感语料和声音事件语料。训练集与六个基准没有共同的录音，也不含评测题特有的文字（附录~\ref{app:g}）。
We trained LoRA adapters in the readout-decline layers, layers 10--25 in Gemma-4 12B and 2--11 in Qwen2.5-Omni-7B, on 103,878 supervised items with 219.0 hours of audio covering the question formats of 34 attribute and structure tasks. The recordings come from 17 corpora, about 80\% of the audio from the LibriSpeech training splits \citep{panayotov2015librispeech} and the rest from Expresso \citep{nguyen2023expresso}, the Speech Accent Archive \citep{weinberger2015accent}, and acted-emotion and sound-event corpora. The training set shares no recordings with the six benchmarks and contains no text specific to their items (Appendix~\ref{app:g}).

% 中文：为了保持基础模型的内容理解准确率，我们加入 16,000 道题的内容锚，包括 10,000 道 LibriSQA 口语理解题 \citep{zhao2023librisqa}，以及 6,000 道基于 LibriSpeech 训练话语和合成训练对话的词类别、情感和讽刺题。对于这些题，我们最小化适配后模型与基础模型在答案位置上的全词表分布之间的 Kullback--Leibler（KL）散度；基础分布在关闭适配器时计算。监督题使用交叉熵，两项合成下面的目标函数。两个模型均取 $\lambda=16$。我们把得到的模型称为修复后的 Gemma-4 12B 和 Qwen2.5-Omni-7B（设置见附录~\ref{app:g}）。
To preserve the base model's accuracy on content questions, we add a content anchor of 16,000 items. It consists of 10,000 LibriSQA spoken-language understanding items \citep{zhao2023librisqa} and 6,000 questions about word categories, emotion and sarcasm built from LibriSpeech training utterances and synthetic training dialogues. On anchor items, we minimize the Kullback--Leibler (KL) divergence between the adapted and base models' vocabulary distributions at answer positions, computing the base distribution with the adapter switched off. Supervised items use cross-entropy. The combined objective is
\[
\mathcal{L}=\mathcal{L}_{\mathrm{CE}}+\lambda\,\mathrm{KL}(p_{\text{base}}\,\|\,p_\theta),
\]
with $\lambda=16$ in both models. We call the results the \emph{repaired} Gemma-4 12B and Qwen2.5-Omni-7B (settings in Appendix~\ref{app:g}).

% 中文：训练后的准确率与录音增益
\subsection{Accuracy and Recording Gain after Training}
\label{sec:results}

% 中文：表~\ref{tab:repair-class} 给出训练前后的类别均值（逐任务结果和按任务组的对照见附录~\ref{app:h}）。属性类准确率在 Gemma-4 12B 中从 34.7% 升至 53.6%，在 Qwen2.5-Omni-7B 中从 56.7% 升至 69.5%，结构类从 35.4% 升至 66.2% 和从 34.0% 升至 78.4%，即属性类提高 18.9 和 12.8 个百分点，结构类提高 30.8 和 44.4 个百分点。平均词汇准确率与基础模型相差不到 1 个百分点，各词汇任务组的变化都在 5 个百分点以内。两个修复后模型的结构类准确率都超过最强开源模型 Qwen3-Omni 的 49.8%，Qwen2.5-Omni-7B 的属性类也超过其 59.9%。与 Gemini 相比，Qwen2.5-Omni-7B 在属性类只差 5.8 个百分点（69.5% 对 75.3%），在结构类则超过 Gemini（78.4% 对 75.4%），而起点排名较低、没有独立编码器的 Gemma-4 12B 仍落后 21.7 和 9.2 个百分点。因此在两个模型中，在读出下降层训练都使属性类和结构类准确率有明显的提高。
Table~\ref{tab:repair-class} gives class means before and after training (per-task results and a task-group view in Appendix~\ref{app:h}). Attribute accuracy rises from 34.7\% to 53.6\% in Gemma-4 12B and from 56.7\% to 69.5\% in Qwen2.5-Omni-7B, and structure accuracy from 35.4\% to 66.2\% and from 34.0\% to 78.4\%, gains of 18.9 and 12.8\,pp on attribute tasks and 30.8 and 44.4\,pp on structure tasks. Mean lexical accuracy stays within 1\,pp of the base model, and every lexical task group within 5\,pp. Both repaired models exceed Qwen3-Omni, the strongest open model, on structure tasks (49.8\%), and Qwen2.5-Omni-7B also on attribute tasks (59.9\%). Against Gemini, Qwen2.5-Omni-7B comes within 5.8\,pp on attribute tasks (69.5\% vs.\ 75.3\%) and exceeds it on structure tasks (78.4\% vs.\ 75.4\%), whereas Gemma-4 12B, which starts from a lower rank and without a separate encoder, remains 21.7 and 9.2\,pp behind. In both models, therefore, training in the readout-decline layers raises attribute and structure accuracy substantially.

% 中文：录音增益随准确率一起上升。属性类的 S/H 在 Gemma-4 12B 中从 −0.4/5.2 升至 12.6/15.7 个百分点，在 Qwen2.5-Omni-7B 中从 18.3/14.9 升至 30.3/28.9 个百分点，结构类从 2.5/5.2 升至 29.6/29.5 和从 1.8/2.7 升至 38.9/46.0 个百分点。在每个类别和模型上，S 的上升占 O 的上升的 69% 至 94%。因此在两种比较下，修复后的模型都从转写之外的录音中获得更多，这一增加构成了提升的大部分。
The recording gains rise with accuracy. On attribute tasks, $S$/$H$ rise from $-0.4$/5.2 to 12.6/15.7\,pp in Gemma-4 12B and from 18.3/14.9 to 30.3/28.9\,pp in Qwen2.5-Omni-7B, and on structure tasks from 2.5/5.2 to 29.6/29.5\,pp and from 1.8/2.7 to 38.9/46.0\,pp. In each class and model, the rise in $S$ accounts for 69\% to 94\% of the rise in $O$. Under both comparisons, therefore, the repaired models obtain more from the recording beyond its transcript, and this increase makes up most of their improvement. 
% 中文：提升从何而来
\subsection{Where the Gains Come From}
\label{sec:repair-readout}

\begin{figure}[t]
\centering
\includegraphics[width=\columnwidth]{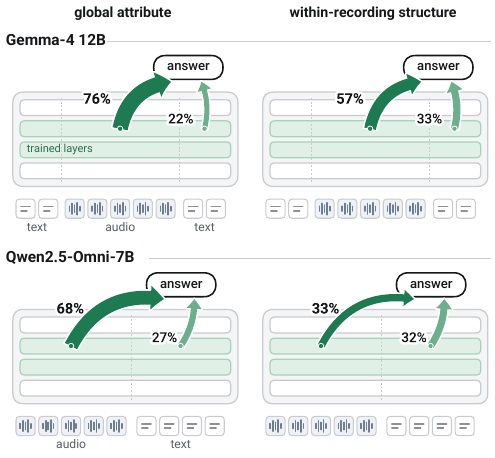}
% 中文：按适配器启用位置分解的准确率提升。箭头粗细表示只在音频位置或只在文本位置启用适配器时，恢复了完整适配器提升的多大份额。
\caption{Accuracy gain by adapter position. Arrow width shows the share of the full adapter's gain recovered when it is active only at audio or text positions.}
\label{fig:repair-route}
\end{figure}

% 中文：训练可能让后面的层保留更多声学信息，也可能让模型更好地利用已经可读的信息。为区分这两种变化，我们在修复后的模型上重复第~\ref{sec:diag} 节的探针测量（附录~\ref{app:f} 的图~\ref{fig:repair-curves}）。在 11 个探针任务上，Gemma-4 12B 和 Qwen2.5-Omni-7B 的作答准确率平均上升 27.7 和 27.2 个百分点，而最后一层的读出只上升 6.7 和 11.0 个百分点。训练前，每个探针任务上的峰值读出都高于作答准确率。训练使这一差距在 Gemma-4 12B 中缩小约一半，在 Qwen2.5-Omni-7B 中缩小大部分。因此，我们把大部分提升归于对表示中已有声学信息的更好利用。读出的变化因任务而异。Gemma-4 12B 性别任务最后一层的读出从 56.6% 升至 72.2%，而说话人计数任务上两个模型的读出变化都不到 5 个百分点，作答准确率却都上升 30 个百分点以上。
Training may improve answers by preserving more acoustic information in later layers or by using information already readable there. To distinguish these changes, we repeated the probe measurements of Section~\ref{sec:diag} on the repaired models (Figure~\ref{fig:repair-curves} in Appendix~\ref{app:f}). Across the 11 probed tasks, answer accuracy rises by 27.7 and 27.2\,pp on average in Gemma-4 12B and Qwen2.5-Omni-7B, whereas final-layer readout rises by only 6.7 and 11.0\,pp. Before training, peak readout exceeded answer accuracy on every probed task. Training closes about half of this difference in Gemma-4 12B and most of it in Qwen2.5-Omni-7B. Because the probes are trained with supervision, this comparison cannot determine how much of the improvement each change explains. The larger rise in answer accuracy than in readout is nevertheless consistent with training improving the use of acoustic information already encoded in the representations. Changes in readout vary by task. For gender in Gemma-4 12B, last-layer readout rises from 56.6\% to 72.2\%, whereas for speaker count readout changes by less than 5\,pp in either model while answer accuracy rises by more than 30\,pp.

% 中文：如果训练主要是让模型学会了题型，适配器的作用应当落在处理题目和答案 token 的文本位置上。为此，我们只在音频位置或只在文本位置（问题、模板与答案 token）启用适配器（图~\ref{fig:repair-route}）。只在音频位置启用时，适配器恢复了属性和结构任务全部提升的 76% 和 57%（Gemma-4 12B）以及 68% 和 33%（Qwen2.5-Omni-7B）。只在文本位置启用时，分别恢复 22% 和 33% 以及 27% 和 32%。在四种情况中的三种，仅音频位置就恢复了一半以上的提升，仅文本位置只恢复 22% 至 33%。例外是 Qwen2.5-Omni-7B 的结构类，仅音频位置和仅文本位置各恢复约三分之一的提升，其余提升需要两处同时启用适配器。因此，提升最大的一块来自训练层对音频 token 处理方式的改变，而这些 token 的表示本来就已携带所需信息（第~\ref{sec:diag} 节），而不是来自读题或写答案方式的改变。
If training had mainly taught the models the question formats, the adapter would act at the text positions, where the question and answer tokens are processed. We therefore applied the adapter only at the audio positions or only at the text positions (question, template and answer tokens) (Figure~\ref{fig:repair-route}). Applied only at the audio positions, the adapter recovers 76\% and 57\% of the full gain on attribute and structure tasks in Gemma-4 12B, and 68\% and 33\% in Qwen2.5-Omni-7B. Applied only at the text positions, it recovers 22\% and 33\%, and 27\% and 32\%. In three of the four cases, the audio positions alone recover more than half of the gain and the text positions alone 22\% to 33\%. The exception is the structure class of Qwen2.5-Omni-7B, where the audio positions alone and the text positions alone each recover about a third of the gain, and the remaining gain requires the adapter at both. The largest part of the improvement therefore comes from changing how the trained layers process the audio tokens, whose representations already carried the information (Section~\ref{sec:diag}), rather than from how the question is read or the answer is written.

\begin{table}[t]
% 中文：两个模型的训练版本对照。$\Delta O$ 和 $\Delta S$ 分别表示各任务类相对基础模型的原录音准确率及 $S$ 的变化，单位为百分点。浅灰行是修复后模型。
\caption{Training variants of Gemma-4 12B and Qwen2.5-Omni-7B. $\Delta O$ and $\Delta S$ give changes from the base model in original-recording accuracy and $S$ by task class (pp), and tinted rows mark the repaired models. We report only $\Delta S$ here, since $S$ and $H$ change in the same direction for both repaired models.}
\label{tab:controls}
\vspace{4pt}
\centering
\footnotesize
\setlength{\tabcolsep}{2.5pt}
\makeatletter % keep the row tint inside the rules: no overhang at the outer edges
\providecommand\tintL{\gdef\CT@cell@color{\@tempdimb\z@\global\let\CT@cell@color\relax}}
\providecommand\tintR{\gdef\CT@cell@color{\@tempdimc\z@\global\let\CT@cell@color\relax}}
\makeatother
\begin{tabular}{@{}l*{5}{>{\raggedleft\arraybackslash}p{2.6em}}@{}}
\toprule
 & \multicolumn{1}{c}{Lexical} & \multicolumn{2}{c}{Attribute} & \multicolumn{2}{c@{}}{Structure} \\
\cmidrule(lr){2-2}\cmidrule(lr){3-4}\cmidrule(l){5-6}
 & \multicolumn{1}{c}{$\Delta O$} & \multicolumn{1}{c}{$\Delta O$} & \multicolumn{1}{c}{$\Delta S$} & \multicolumn{1}{c}{$\Delta O$} & \multicolumn{1}{c@{}}{$\Delta S$} \\
\midrule
\multicolumn{6}{@{}l}{\textbf{Gemma-4 12B}} \\
\rowcolor{black!8}
\tintL Layers 10--25 & $+0.8$ & $+18.9$ & $+13.0$ & $+30.8$ & \tintR$+27.0$ \\
No content anchor & $-10.5$ & $+16.5$ & $+15.1$ & $+35.2$ & $+29.4$ \\
Layers 0--15 & $+1.1$ & $+20.3$ & $+13.7$ & $+16.3$ & $+11.2$ \\
Layers 32--47 & $+0.2$ & $+8.2$ & $+3.9$ & $+11.4$ & $+7.6$ \\
All 48 layers & $-2.5$ & $+17.2$ & $+14.7$ & $+34.9$ & $+26.8$ \\
\midrule
\multicolumn{6}{@{}l}{\textbf{Qwen2.5-Omni-7B}} \\
\rowcolor{black!8}
\tintL Layers 2--11 & $+0.3$ & $+12.8$ & $+12.0$ & $+44.4$ & \tintR$+37.2$ \\
No content anchor & $-5.6$ & $+13.6$ & $+14.3$ & $+49.6$ & $+42.1$ \\
Layers 12--21 & $-1.4$ & $+9.9$ & $+9.9$ & $+30.1$ & $+22.2$ \\
Layers 18--27 & $-1.0$ & $+7.8$ & $+7.3$ & $+19.2$ & $+13.3$ \\
All 28 layers & $+0.7$ & $+13.8$ & $+13.9$ & $+44.4$ & $+34.7$ \\
\bottomrule
\end{tabular}
\end{table}

% 中文：消融与迁移
\subsection{Ablations and Transfer}
\label{sec:placement}

% 中文：我们接着考察训练选择对这些结果的影响。表~\ref{tab:controls} 中的每个变体相对修复后模型只改动一项选择。去掉内容锚后，两个模型的词汇准确率都下降，而属性类和结构类准确率的提升相近或更大，$\Delta S$ 也不更小。因此，内容锚保持了词汇准确率，同时保留了无锚点时 $\Delta S$ 的 84--92\%。把适配器放到其他等宽层段后，两个模型的结构类提升都减小，靠后的层段减小最多，$\Delta S$ 随之减小。训练全部层所用的可训练参数约为三倍，属性类和结构类的提升与读出下降层相差至多 4.1 个百分点，$\Delta S$ 相差至多 2.5 个百分点。因此，提升取决于适配器放在哪里，而不是训练了多少参数。靠后的层段中读出已经下降（图~\ref{fig:probe}），恢复的提升更少，而读出下降层之外的参数贡献很小。
We next examine how the training choices affect these results. Each variant in Table~\ref{tab:controls} changes one choice relative to the repaired model. Without the content anchor, lexical accuracy falls in both models, whereas attribute and structure accuracy rise by similar or larger amounts and $\Delta S$ is no smaller. The anchor therefore preserves lexical accuracy while retaining 84--92\% of the $\Delta S$ obtained without it. Placing the adapters in other ranges of equal width reduces the structure gain in both models, most for the later ranges, and $\Delta S$ decreases with it. Training all layers, with about three times as many trainable parameters, changes the attribute and structure gains by at most 4.1\,pp and $\Delta S$ by at most 2.5\,pp relative to the readout-decline layers. The gains therefore depend on where the adapter is placed rather than on how many parameters are trained. The later ranges, in which readout has already declined (Figure~\ref{fig:probe}), recover less of the gain, and parameters outside the readout-decline layers contribute little.

% 中文：为检验向训练中没有的题型的迁移，我们去掉全部性别和词重音训练题，另外训练了一个 Gemma-4 12B 模型。这两类任务的准确率仍然上升，$S$ 的增加约为完整训练的六成。其余训练任务涉及相关的线索与操作，可能解释这一迁移。性别、说话人和年龄题使用部分相同的嗓音线索，包括音高。词重音题需要定位一个词并比较其韵律；词时间戳题也需要定位词，音高和时长比较题则需要比较声学属性。在训练中没有同题型样本的七项对话情感和讽刺任务上，完整修复后的 Gemma-4 12B 的原录音准确率平均下降 4.1 个百分点，而 Qwen2.5-Omni-7B 平均提高 4.3 个百分点。这些结果表明，共享线索和操作的任务之间可能发生迁移，但未见题型上的效果并不一致。
Finally, to test transfer to formats absent from training, we trained a separate Gemma-4 12B model without gender or word-stress items. Accuracy on both task types still rises, and the increase in $S$ is about 60\% of that after full training. Other training tasks involve related cues and operations that may explain this transfer. Gender, speaker, and age questions share some voice cues, including pitch. Word-stress questions require locating a word and comparing its prosody. Word-timestamp questions also require locating words, while pitch- and duration-comparison questions require comparing acoustic properties. Across the seven conversational emotion and sarcasm tasks without matching training formats, original-recording accuracy falls by 4.1\,pp on average in the fully repaired Gemma-4 12B but rises by 4.3\,pp in Qwen2.5-Omni-7B. These results suggest transfer between tasks that share cues and operations, but the gains on unseen formats are uneven.

%% file: sec/06_related.tex
% 中文：相关工作
\section{Related Work}
\label{sec:related}

% 中文：要估计录音在转写之外的贡献，已有工作需要保留转写中的词作为参照。加噪、用静音替换或去除音频同样会削弱词汇证据 \citep{sakshi2024mmau,wang2026mmsu,he2026measuring,foo2026glitters}。以文字为参照保留了词，却把输入从语音换成了文字 \citep{cuervo2026closing}。另一类研究让声音与说出的词或附加的文字相冲突，观察模型跟随哪一方 \citep{chen2025listen,xiong2026deaf,wang2025mcr,pang2026voxparadox,billa2026alme}。ALME \citep{billa2026alme} 在保留词汇内容的前提下重新合成刺激，最接近我们的参照设计。我们把同转写的合成朗读配给现有基准中的自然录音，比较的两侧都是语音，同一批朗读可用于比较不同模型以及同一模型训练前后的 S，并以 H 作为补充。
To estimate what a recording contributes beyond its transcript, we need a reference that retains the transcribed words. Adding noise, replacing recordings with silence or removing audio weakens the lexical evidence as well \citep{sakshi2024mmau,wang2026mmsu,he2026measuring,foo2026glitters}, and a text reference preserves the words but changes the input from speech to text \citep{cuervo2026closing}. Other work places the voice in conflict with the spoken words or with added text to see which source the model follows \citep{chen2025listen,xiong2026deaf,wang2025mcr,pang2026voxparadox,billa2026alme}. ALME \citep{billa2026alme}, closest to our reference design, resynthesizes stimuli with the same lexical content. We pair natural recordings with synthetic readings of their transcripts, so that both inputs remain speech and the same readings serve to compare $S$ across models and before and after training, and we complement this comparison with $H$.

% 中文：我们要问的是，模型没用上的信息是模型里没有，还是在表示中有而没用上。已有研究从隐藏状态中读出了物理属性、韵律差异或音频支持的答案，模型自身的作答却接近随机，或跟随冲突的文字 \citep{sun2026sonicbench,wang2025mcr,gao2026arbitration,peng2026represented}。VoxParadox \citep{pang2026voxparadox} 将评测、探针与训练修复结合，与本文最接近，它在 Audio Flamingo 3 \citep{goel2025af3} 中发现说话人线索在语言模型中仍可读，而比较线索在编码器深层和投影处减弱。这些音频研究都考察带独立编码器的模型。已有工作报告无编码器设计提高了副语言判断准确率 \citep{fan2026melllm}，我们则发现无编码器的 Gemma-4 12B 同样存在利用缺口，并在读出下降的层段训练来检验作答能否改善。
We ask whether the information a model leaves unused is absent from it or present in its representations. Prior studies recover physical attributes, prosodic contrasts or audio-supported answers from hidden states while the models answer near chance or follow conflicting text \citep{sun2026sonicbench,wang2025mcr,gao2026arbitration,peng2026represented}. VoxParadox \citep{pang2026voxparadox}, closest to our work, combines evaluation, probing and repair and finds in Audio Flamingo 3 \citep{goel2025af3} that speaker cues remain readable in the language model while comparison cues weaken in deep encoder layers and at the projection. These audio studies examine models with separate encoders. Although encoder-free designs are reported to improve paralinguistic accuracy \citep{fan2026melllm}, we find the same utilization gap in encoder-free Gemma-4 12B and test whether training where readout declines improves answers.

% 中文：要训练模型使用这些信息，还需要检查模型是否保留了对口语内容的理解。已有方法在解码时根据音频或对比规则调整输出分数 \citep{hsu2025audioaware,lin2026contrastive,gao2026arbitration}，或按提示混合编码器各层表示后再做直接偏好优化 \citep{pang2026voxparadox,rafailov2023direct}。在音频与文字冲突的样本上微调，可能降低两者一致或只有音频时的准确率 \citep{wang2025mcr,billa2026alme}，因此需要单独评估原有能力的保持 \citep{lu2025speechifeval}。已有做法用原模型的输出约束训练，或用原模型、骨干模型给出的目标训练语音模型 \citep{li2016learning,held2024diva,lu2025desta}。我们沿用这一思路，在读出下降层训练 LoRA 适配器，并用内容锚使内容题的回答贴近基础模型。
To train the model to use this information, we also need to check that it retains its understanding of spoken content. Existing methods adjust output scores at decoding time with audio-aware or contrastive rules \citep{hsu2025audioaware,lin2026contrastive,gao2026arbitration}, or mix encoder layers according to the prompt before direct preference optimization \citep{pang2026voxparadox,rafailov2023direct}. Fine-tuning on conflicting audio and text can lower accuracy when the inputs agree or only audio is given \citep{wang2025mcr,billa2026alme}, and retention therefore needs its own evaluation \citep{lu2025speechifeval}. Established approaches constrain training with the original model's outputs or train speech models on targets from the original or backbone model \citep{li2016learning,held2024diva,lu2025desta}. We build on this idea by training LoRA adapters in the readout-decline layers with a content anchor that keeps answers to content questions close to the base model's.

%% file: sec/07_conclusion.tex
% 中文：结论
\section{Conclusion}
\label{sec:conclusion}

% 中文：我们在六个基准的 92 个任务上衡量了原录音在词之外对作答的贡献。在属性类和结构类任务上，录音增益之差占 19 个开源模型均值与 Gemini 3.1 Pro 准确率差距的一半以上。在 Gemma-4 12B 和 Qwen2.5-Omni-7B 中，即使模型在同一批题上作答较差，线性探针仍能从语言模型早期层读出相关声学线索。在我们构造的数据上、于读出下降的层训练 LoRA 适配器并加入内容锚后，两个模型的属性类和结构类准确率提高 12.8 至 44.4 个百分点。大部分提升体现为录音增益增大，平均词汇准确率与基础模型相差不到 1 个百分点。在探针任务上，作答准确率的提高也大于读出的变化，支持模型更好地利用表示中已有的信息。因此，我们研究的开源模型编码的声学信息多于它们用到的，而训练改变了它们处理这些信息的方式。
We estimated the contribution of the original recording beyond its words across 92 tasks from six benchmarks. On attribute and structure tasks, differences in recording gain account for about half or more of the accuracy gap between the mean of 19 open models and Gemini 3.1 Pro. In Gemma-4 12B and Qwen2.5-Omni-7B, linear probes read out relevant acoustic cues from early language-model layers even when the models answer the same items poorly. Training LoRA adapters on our constructed data in the layers where readout declines, with a content anchor, raises attribute and structure accuracy by 12.8 to 44.4\,pp in the two models. Most of this increase appears as larger recording gains, while mean lexical accuracy stays within 1\,pp of the base models. On the probed tasks, answer accuracy also rises more than readout, consistent with better use of information already encoded in the representations. The two probed models therefore encode acoustic information that they do not reliably use in answers, and training changes how they process it.

%% file: sec/08_limitations.tex
% 中文：局限
\section*{Limitations}
\label{sec:limitations}

% 中文：我们使用两种录音增益 来 衡量原录音相对于不同参照的准确率优势。$S$ 以合成朗读为参照，因此也包含自然语音与合成语音之间的差异。合成声音的属性可能帮助或误导模型。$H$ 以转写为参照，而提供转写可能改变模型结合文字与音频的方式。尽管我们使用两种方式来估计增益，提供了鲁棒的估计。两种增益呈现一致规律，支持模型利用录音的程度存在差异。但是两种比较都依赖同一份 Whisper 转写，识别错误或遗漏可能同时影响两者。
\paragraph{Reference conditions.}
We use two recording gains to estimate the original recording's accuracy advantage relative to different references. $S$ uses synthetic readings and thus includes differences between natural and synthetic speech. Properties of the synthetic voices may help or mislead the model. $H$ uses the transcript as its reference, and providing it may change how the model combines text and audio. Estimating the gain in two ways yields a more robust estimate. The two gains show a consistent pattern, which supports differences in how much models use the recording. However, both comparisons depend on the same Whisper transcript, and recognition errors or omissions can therefore affect them together.

% 中文： 我们只在两个模型上检验了修复，一个有独立音频编码器，另一个没有。两者的规模和训练数据也不同，因此它们之间的差别不能归因于编码器设计。例如，训练前 Qwen2.5-Omni-7B 的属性类 $S$/$H$ 高于 Gemma-4 12B（18.3/14.9 对 −0.4/5.2 个百分点），训练后补回的读出与作答准确率差距的比例也更大。还需在更多架构上检验修复能推广到多大范围。我们猜测这两个架构的模型可能存在有趣的区别，但是本文没有深入探索。
\paragraph{Number of repaired models.}
We tested the repair in only two models, one with and one without a separate audio encoder. They also differ in size and training data, so differences between them cannot be attributed to encoder design. Before training, for example, Qwen2.5-Omni-7B has higher attribute-class $S$/$H$ than Gemma-4 12B (18.3/14.9 vs.\ $-0.4$/5.2\,pp), and after training it closes a larger share of the gap between readout and answer accuracy. Testing further architectures is needed to establish how broadly the repair applies. We suspect that models with these two architectures differ in further notable ways, which this paper does not examine in depth.

% 中文：【迁移与副作用】训练题的题型与 34 个评测任务对齐，因此超出这些题型的迁移仍需检验。在没有同题型训练样本的七项对话情感和讽刺任务上，Gemma-4 12B 的平均准确率下降 4.1 个百分点，Qwen2.5-Omni-7B 则提高 4.3 个百分点，而且两个模型都有个别任务下降。另行训练的 Gemma-4 12B 去掉了性别和词重音题，结果显示这两类判断仍有提升。这些不一致的结果说明迁移能力还需要更加仔细的分析。
\paragraph{Transfer and side effects.}
The training items match the question formats of 34 evaluation tasks, so transfer beyond these formats needs further testing. Across the seven conversational emotion and sarcasm tasks without matching training formats, mean accuracy falls by 4.1\,pp in Gemma-4 12B but rises by 4.3\,pp in Qwen2.5-Omni-7B, with declines on individual tasks in both models. In a separate Gemma-4 12B run without gender or word-stress training items, both judgements still improve. These mixed results call for more careful analysis of transfer.

% 中文：【探针读出】探针在各任务标签上经过有监督训练，在结构题上还获得片段边界，而模型作答时两者都没有。因此，读出与作答准确率之间的差距也反映了这些优势，单凭这一差距不能说明模型没有利用已有信息。同样，训练后作答准确率的上升大于读出的上升，也不能定量说明多少提升来自对已有信息的更好利用。
\paragraph{Probe readout.}
Probes are trained with supervision on each task's labels and receive the segment boundaries on structure items, whereas the models answer without either. The gap between readout and answer accuracy therefore also reflects these advantages and does not by itself show that the models leave available information unused. Likewise, the larger rise in answer accuracy than in readout after training does not quantify how much of the improvement comes from better use of existing information.

% 中文：【修复作用的位置】读出下降引导了层段选择，但读出保持较高时作答也可能较差，因此读出不能单独确定最有效的训练层段。训练全部层时，可训练参数约为三倍，$\Delta S$ 与只训练选定层段相比至多相差 2.5 个百分点。每个模型的层段对照只比较了另外两个等宽范围，更细的搜索可能找到增益更大的位置。只在音频位置或只在文本位置启用适配器的拆分，表明在这一设置下，音频位置的改动有助于提升。具体改变了哪部分内部计算，还需要进一步的干预实验。
\paragraph{Where the repair acts.}
Readout decline guided layer selection, but answers can also be poor where readout remains high, so readout alone cannot establish the most effective training range. Training all layers uses about three times as many trainable parameters and changes $\Delta S$ by at most 2.5\,pp relative to the selected layers. The layer controls compare only two other equal-width ranges per model, and a finer search could find larger gains. Enabling the adapters only at audio positions or only at text positions shows that, in this setting, changes at audio positions contribute to the gains. Identifying which part of the internal computation changes requires further intervention experiments.

% 中文：【评测设置】多数题目只要求模型回答关于一段录音的一个问题，涉及多人对话的任务只占一小部分。全双工语音模型边听边说，需要在对方尚未说完时，根据语调、停顿和情绪等线索决定何时接话、如何回应。它们在这种连续交互中是否利用词之外的声音，以及合成朗读是否仍是衡量录音贡献的有用参照，都需要另行检验。
\paragraph{Evaluation setting.}
Most items ask one question about one recording, and tasks involving multi-party dialogue form only a small part of the evaluation. Full-duplex speech models listen while speaking and must use cues such as intonation, pauses and emotion before the other speaker has finished to decide when and how to respond. Whether they use sound beyond the words in such continuous interaction, and whether synthetic readings remain a useful reference for measuring the recording's contribution, require separate tests.

%% file: sec/app_a_inputs.tex
% 中文：数据与输入
\section{Data}
\label{app:a}

% 中文：同一道题在五种输入下使用相同的问题、选项和判分规则。因此，我们只保留六个公开基准中题目以文字给出、答案可以自动判分的部分（表~\ref{tab:a1}）。AudioBench、AIR-Bench 和 Dynamic-SUPERB 每个任务分别至多取 500、150 和 200 道题，其余三个基准全部保留。录音包括表演、影视对白和访谈，也包括朗读语音与合成语音。
Each item uses the same question, options and scoring rule under all five inputs. We therefore retained only the parts of six public benchmarks with written questions and automatically scorable answers (Table~\ref{tab:a1}). We took at most 500, 150 and 200 items per task from AudioBench, AIR-Bench and Dynamic-SUPERB, respectively, and kept all items from the other three benchmarks. The recordings include acted speech, film and television dialogue, interviews, read speech and synthetic speech.

\begin{table*}[t]
% 中文：六个基准中使用的数据。表中列出取用部分、答案形式、任务数和题数，共 13,707 道题。
\caption{Benchmark subsets used in the study. The table lists their answer formats, task counts and item counts (13{,}707 items in total).}
\label{tab:a1}
\centering
\small
\begin{tabularx}{\linewidth}{@{}l>{\raggedright\arraybackslash}X>{\raggedright\arraybackslash}p{0.25\linewidth}rr@{}}
\toprule
Benchmark & Part used & Answer format & Tasks & Items \\
\midrule
AudioBench & Voice understanding (6 sets), spoken QA (4 sets) & Label, multiple choice, short answer & 10 & 4{,}779 \\
AIR-Bench & Foundation, speech tasks & Multiple choice & 8 & 1{,}161 \\
Dynamic-SUPERB Phase-2 & 11 speech tasks & Label & 11 & 2{,}035 \\
MMAU & test-mini, speech & Multiple choice & 10 & 332 \\
MMAU-Pro & Speech, one recording per item & Multiple choice & 6 & 865 \\
MMSU & All tasks & Multiple choice & 47 & 4{,}535 \\
\midrule
Total & & & 92 & 13{,}707 \\
\bottomrule
\end{tabularx}
\end{table*}

% 中文：选择题按回答选中的选项判分，标签题按回答最先给出的标签判分。AudioBench 的两个问答子集按回答与标准答案的词重叠判分。同一道题在五种输入下沿用同一规则。Whisper-large-v3 \citep{radford2022whisper} 生成转写，$A$ 给出转写文字，$T$ 给出它的合成朗读。Mustard 讽刺检测任务是例外，大多数朗读使用数据集自带的人工转写。Edge-TTS \citep{edgetts} 用女声 T1 和男声 T2 各朗读一次。我们再用 Whisper 转写这些朗读，剔除词错率超过 20\% 的版本。保留下来的朗读相对源转写的词错率为 1.4\%（第~\ref{sec:design} 节）。
We score multiple-choice items by the selected option and label items by the first label in the response. Each item keeps the same scoring rule across all five inputs. Whisper-large-v3 \citep{radford2022whisper} produces the transcript used as text in $A$ and as a synthetic reading in $T$. For Mustard sarcasm detection, most readings instead use the dataset's human transcript. Edge-TTS \citep{edgetts} reads each transcript in a female voice (T1) and a male voice (T2). We transcribe these readings again with Whisper and re-synthesize the readings  until the word error rate is below 20\%. The retained readings have a word error rate of 1.4\% relative to their source transcripts (Section~\ref{sec:design}).

%% file: sec/app_b_classes.tex
% 中文：任务分类
\section{Task Classification}
\label{app:b}

% 中文：任务类别取决于作答所需的证据（表~\ref{tab:b2}）。dream\_tts\_mcq\_test 问对话内容，属于词汇类。Speaker\_Gender\_Recognition 问说话人性别，属于全局属性类。名为 pitch\_comparison 的任务比较两段的音高，属于录音内部结构类。混合类同时需要声音和词。例如，stress\_based\_reasoning 要先听出哪个词有重音，再结合句意判断它强调了什么。
We classify tasks by the evidence needed for an answer (Table~\ref{tab:b2}). The dialogue-content task dream\_tts\_mcq\_test is lexical. Speaker\_Gender\_Recognition asks for the speaker's gender and is a global-attribute task. The pitch\_comparison task compares two segments and belongs to within-recording structure. Mixed tasks require both sound and words. In stress\_based\_reasoning, for example, the model must identify the stressed word and use the sentence's meaning to interpret the emphasis.

% 中文：我们排除了 Dynamic-SUPERB 的 StressDetection\_MIRSD（173 道题）。任务描述问录音中的重音，标准答案却是孤立单词的词典重音，通常仅凭文字就能判断。只给转写时，Gemini 在这一任务的 40 道题上答对 92.5\%，因此用它评测录音内部结构会混入词汇线索。
We excluded StressDetection\_MIRSD from Dynamic-SUPERB (173 items). Its description asks about stress in the recording, but its gold labels give the dictionary stress of isolated words, which can usually be determined from text. Given only the transcript, Gemini answers 92.5\% of 40 items correctly, so treating this task as within-recording structure would introduce a lexical shortcut.

\begin{table*}[t]
% 中文：按证据类别和基准列出的任务。任务名沿用原基准，括号内为任务数。
\caption{Tasks by evidence class and benchmark. Names follow the source benchmarks, and parentheses give task counts.}
\label{tab:b2}
\centering
\footnotesize
\begin{tabularx}{\linewidth}{@{}l>{\raggedright\arraybackslash}X@{}}
\toprule
Benchmark & Tasks \\
\midrule
\multicolumn{2}{@{}l}{\emph{Lexical (36 tasks)}} \\
AIR-Bench (2) & Speaker\_Intent\_Classification, Speech\_Entity\_Reconition \\
AudioBench (4) & cn\_college\_listen\_mcq\_test, dream\_tts\_mcq\_test, public\_sg\_speech\_qa\_test, slue\_p2\_sqa5\_test \\
Dynamic-SUPERB (2) & DialogueActClassification\_DailyTalk, SuperbIC\_SLURP-Intent \\
MMAU (7) & Conversational Fact Retrieval, Dissonant Emotion Interpretation, Event-Based Knowledge Retrieval, Key highlight Extraction, Multi Speaker Role Mapping, Phonemic Stress Pattern Analysis, Phonological Sequence Decoding \\
MMSU (21) & casual\_reasoning, code\_switch\_question\_answering, consonant\_vowel\_perception, content\_grounding, continuation\_writing, couplet\_matching, deixis\_resolution, dialogue\_reasoning, homophone\_based\_reasoning, idiom\_reasoning, intent\_detection, logical\_reasoning, long\_speech\_summarization, near\_homophone\_perception, polysemy\_reasoning, prolonged\_sound\_reasoning, puns\_interpretation, speech\_act\_classification, speech\_translation, syllable\_perception, syntactic\_structure\_matching \\
\midrule
\multicolumn{2}{@{}l}{\emph{Global attribute (30 tasks)}} \\
AIR-Bench (4) & Speaker\_Age\_Prediction, Speaker\_Emotion\_Recontion, Speaker\_Gender\_Recognition, Speaker\_Number\_Verification \\
AudioBench (6) & iemocap\_emotion\_recognition, iemocap\_gender\_recognition, meld\_emotion\_test, meld\_sentiment\_test, voxceleb\_accent\_test, voxceleb\_gender\_test \\
Dynamic-SUPERB (8) & AccentClassification\_AccentdbExtended, DialogueEmotionClassification\_DailyTalk, EmotionRecognition\_MultimodalEmotionlinesDataset, MultiSpeakerDetection\_VCTK, NoiseDetection\_VCTK\_MUSAN-Noise, ReverberationDetection\_VCTK\_RirsNoises-MediumRoom, SarcasmDetection\_Mustard, SpeakerCounting\_LibriTTS-TestClean \\
MMAU (2) & Counting, Emotion Flip Detection \\
MMSU (10) & accent\_identification, age\_prediction, background\_scene\_recognition, dialogue\_turn\_counting, emotion\_recognition, emotional\_context\_reasoning, gender\_prediction, non\_verbal\_sound\_detection, sarcasm\_detection, total\_speaker\_counting \\
\midrule
\multicolumn{2}{@{}l}{\emph{Within-recording structure (12 tasks)}} \\
AIR-Bench (1) & Speech\_Grounding \\
MMSU (11) & disfluency\_detection, intonation\_perception, pause\_perception, pitch\_comparison, plosive\_sound\_identification, prolonged\_sound\_perception, speaker\_identity\_recognition, speech\_duration\_estimation, speech\_stress\_perception, speed\_comparison, volume\_comparison \\
\midrule
\multicolumn{2}{@{}l}{\emph{Mixed (14 tasks)}} \\
AIR-Bench (1) & Synthesized\_Voice\_Detection \\
Dynamic-SUPERB (1) & SpoofDetection\_ASVspoof2017 \\
MMAU (1) & Emotion State summarisation \\
MMAU-Pro (6) & Audio Quality, Artifacts \& Channel Characteristics; Lexical and Phrase-Level Recognition; Paralinguistic/Emotion Recognition; Speaker Demographics; Speaker Identification; Speech Activity, Turn-Taking and Overlap Detection \\
MMSU (5) & intonation\_based\_reasoning, pause\_based\_reasoning, stress\_based\_reasoning, synthetic\_speech\_detection, vocal\_range\_comparison \\
\bottomrule
\end{tabularx}
\end{table*}

%% file: sec/app_c_systems.tex
% 中文：评测系统
\section{Systems}
\label{app:c}

% 中文：表~\ref{tab:c1} 列出 19 个开源模型及实际使用的检查点。闭源参照模型是 Gemini 3.1 Pro。
Table~\ref{tab:c1} identifies the checkpoints used for the 19 open models. Gemini 3.1 Pro is the closed-model reference.

\begin{table*}[t]
% 中文：本文评测的开源端到端音频模型及检查点。
\caption{Open end-to-end audio models and evaluated checkpoints.}
\label{tab:c1}
\centering
\footnotesize
\setlength{\tabcolsep}{4pt}
\begin{tabular}{@{}lll@{}}
\toprule
Model & Reference & Checkpoint \\
\midrule
Qwen3-Omni & \citealp{xu2025qwen3omni} & \texttt{Qwen/Qwen3-Omni-30B-A3B-Instruct} \\
Qwen2.5-Omni-7B & \citealp{xu2025qwen25omni} & \texttt{Qwen/Qwen2.5-Omni-7B} \\
MiniCPM-o 4.5 & \citealp{cui2026minicpmo} & \texttt{openbmb/MiniCPM-o-4\_5} \\
Qwen2.5-Omni-3B & \citealp{xu2025qwen25omni} & \texttt{Qwen/Qwen2.5-Omni-3B} \\
MiniCPM-o 2.6 & \citealp{yao2024minicpmv} & \texttt{openbmb/MiniCPM-o-2\_6} \\
Kimi-Audio & \citealp{kimiteam2025kimiaudio} & \texttt{moonshotai/Kimi-Audio-7B-Instruct} \\
Audio Flamingo 3 & \citealp{goel2025af3} & \texttt{nvidia/audio-flamingo-3-hf} \\
Audio Flamingo Next & \citealp{ghosh2026afnext} & \texttt{nvidia/audio-flamingo-next-hf} \\
MERaLiON-3 & \citealp{he2024meralion} & \texttt{MERaLiON/MERaLiON-3-10B} \\
Step-Audio 2 & \citealp{wu2025stepaudio2} & \texttt{stepfun-ai/Step-Audio-2-mini} \\
Voxtral-Mini-3B & \citealp{liu2025voxtral} & \texttt{mistralai/Voxtral-Mini-3B-2507} \\
MERaLiON-2 & \citealp{he2024meralion} & \texttt{MERaLiON/MERaLiON-2-10B} \\
Gemma-4 E4B & \citealp{gemmateam2026gemma4} & \texttt{google/gemma-4-E4B-it} \\
Qwen2-Audio & \citealp{chu2024qwen2audio} & \texttt{Qwen/Qwen2-Audio-7B-Instruct} \\
R1-AQA & \citealp{li2025r1aqa} & \texttt{mispeech/r1-aqa} \\
Phi-4-Multimodal & \citealp{abouelenin2025phi4mini} & \texttt{microsoft/Phi-4-multimodal-instruct} \\
Aero-1-Audio & \citealp{lmmslab2025aero} & \texttt{lmms-lab/Aero-1-Audio} \\
Gemma-4 E2B & \citealp{gemmateam2026gemma4} & \texttt{google/gemma-4-E2B-it} \\
Gemma-4 12B & \citealp{gemmateam2026gemma4} & \texttt{google/gemma-4-12B-it} \\
\bottomrule
\end{tabular}
\end{table*}

%% file: sec/app_e_readout_method.tex
% 中文：读出与作答的测量
\section{Measuring Readout and Answers}
\label{app:e}

% 中文：线性探针需要声学属性已知的题。属性类任务可以直接使用评测题。性别探针合并 AudioBench 的 880 道 IEMOCAP 和 VoxCeleb 性别题，其他属性探针使用 30 个任务中题量足够的 26 个，每个任务有 100 至 308 道题。结构类任务的原题没有标明目标声音的位置，因此我们为其中 12 个任务中的 5 个另外构造录音，分别考察音高、语速和响度比较，以及词重音和延长音。
Linear probes require items with known acoustic labels. For attribute tasks, we use the evaluation items. The gender probe combines 880 IEMOCAP and VoxCeleb items from AudioBench, while the other probes use 26 of the 30 attribute tasks with sufficient data (100--308 items per task). The original structure items do not identify the location of the relevant sound, so we constructed recordings for five of the 12 tasks. These cover pitch, speaking-rate and loudness comparison, word stress, and prolonged sound.

% 中文：三个比较任务均将同一说话人的两段 LibriSpeech 语句用 0.5 秒静音连接，再让两段在目标属性上朝相反方向变化。音高差分别为 1.5、2、4 和 8 个半音，语速比为 1.2、1.5 和 2.0，响度差为 3、6 和 12 dB，每个档位有 200 段录音。最小的音高差 1.5 个半音，接近 MMSU 音高比较相邻档位的 1.66 和 1.82 个半音差。词重音题同时抬高一个词的音高（中位数 4.1 个半音）并延长它，共 600 道；延长音题只延长该词，共 585 道，延长比例中位数为 2.38。
For the three comparison tasks, we joined two LibriSpeech utterances \citep{panayotov2015librispeech} from the same speaker with 0.5~s of silence, then changed the target property in opposite directions. Pitch gaps were 1.5, 2, 4, and 8 semitones. Speaking-rate ratios were 1.2, 1.5, and 2.0, and loudness gaps were 3, 6, and 12~dB. We made 200 recordings at each level. The smallest pitch gap, 1.5 semitones, is close to the 1.66 and 1.82 semitone gaps between adjacent levels in MMSU pitch comparison. For word stress, we raised the pitch of one word (by a median of 4.1 semitones) and lengthened it, producing 600 items. For prolonged sound, we only lengthened the word, producing 585 items with a median lengthening ratio of 2.38.

% 中文：每个逻辑回归探针使用一层音频位置上标准化后的表示训练，并在训练中未出现的题上测试。例如，性别探针在一个数据集上训练、另一个数据集上测试。构造的结构题按说话人分组做五折交叉验证，使同一说话人不会同时出现在训练和测试中。作答准确率则来自模型对原属性题的回答，或对构造结构题的选项字母选择。
We train each logistic regression probe on standardized representations at one layer's audio positions and test it on items held out from its training data. For example, the gender probe is trained on one dataset and tested on the other. For constructed structure items, we use five-fold cross-validation grouped by speaker, which keeps the same speaker out of both training and test data. Answer accuracy comes from the model's response to the original attribute question or its choice of option letter on a constructed structure item.

%% file: sec/app_f_readout_results.tex
% 中文：全部探针任务的逐层读出
\section{Layer-wise Readout on All Probed Tasks}
\label{app:f}

\begin{figure*}[tp]
\centering
\includegraphics[width=\linewidth]{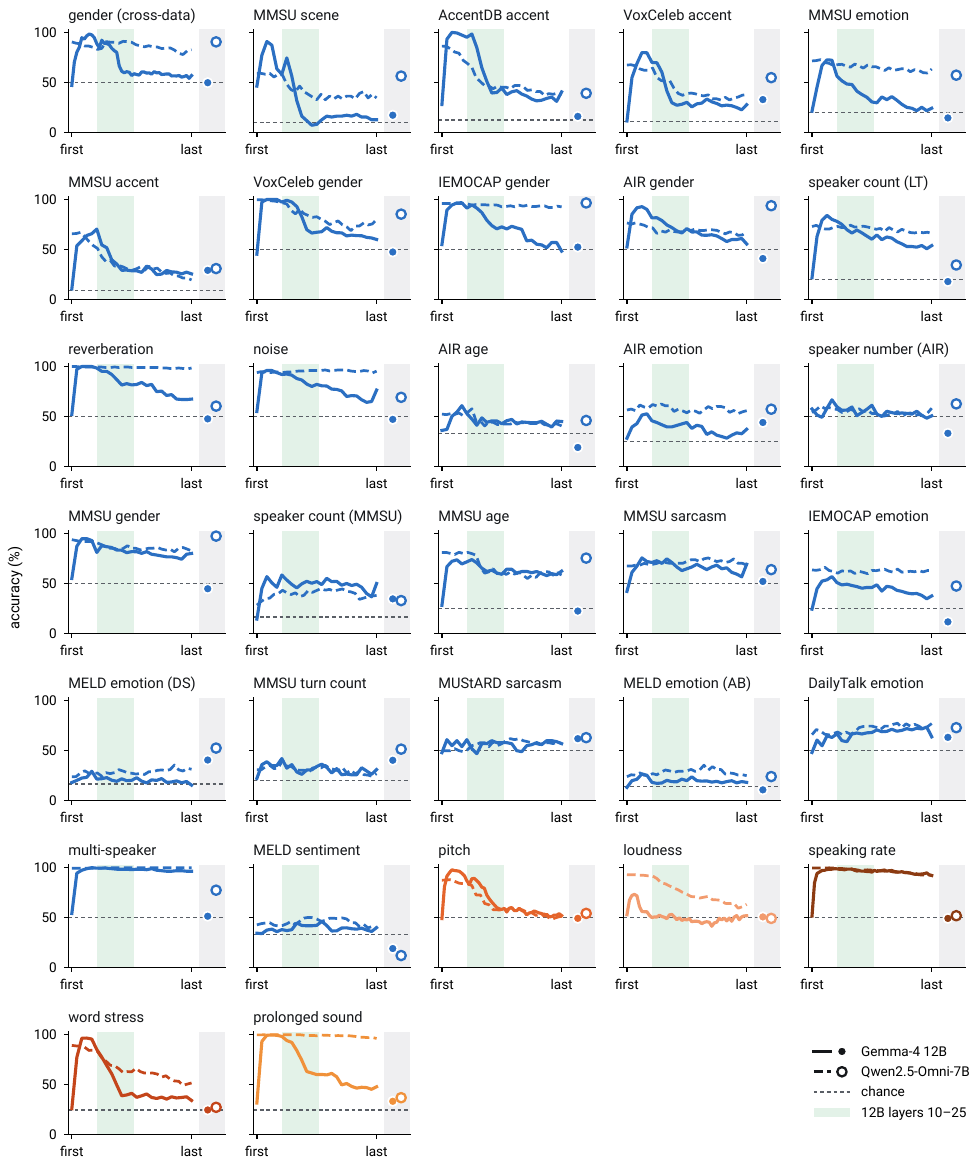}
% 中文：所有探针任务的逐层准确率。蓝色面板为属性题，橙色面板为构造的结构题；实线和实心点表示 Gemma-4 12B，虚线和空心点表示 Qwen2.5-Omni-7B，右侧点为原录音作答准确率，灰色虚线为随机水平，绿色带为 Gemma-4 12B 第 10–25 层；LT、DS、AB、AIR 分别指 LibriTTS、Dynamic-SUPERB、AudioBench、AIR-Bench。
\caption{Probe accuracy across relative layer depth on all tasks. Blue panels show attributes and orange panels constructed structure items, with solid lines and filled answer dots for Gemma-4 12B, dashed lines and open dots for Qwen2.5-Omni-7B, a grey dashed chance line, and a green band over Gemma-4 12B layers 10--25 (LT, DS, AB, AIR denote LibriTTS, Dynamic-SUPERB, AudioBench, AIR-Bench).}
\label{fig:probe-all}
\end{figure*}

\begin{figure*}[tp]
\centering
\includegraphics[width=\linewidth]{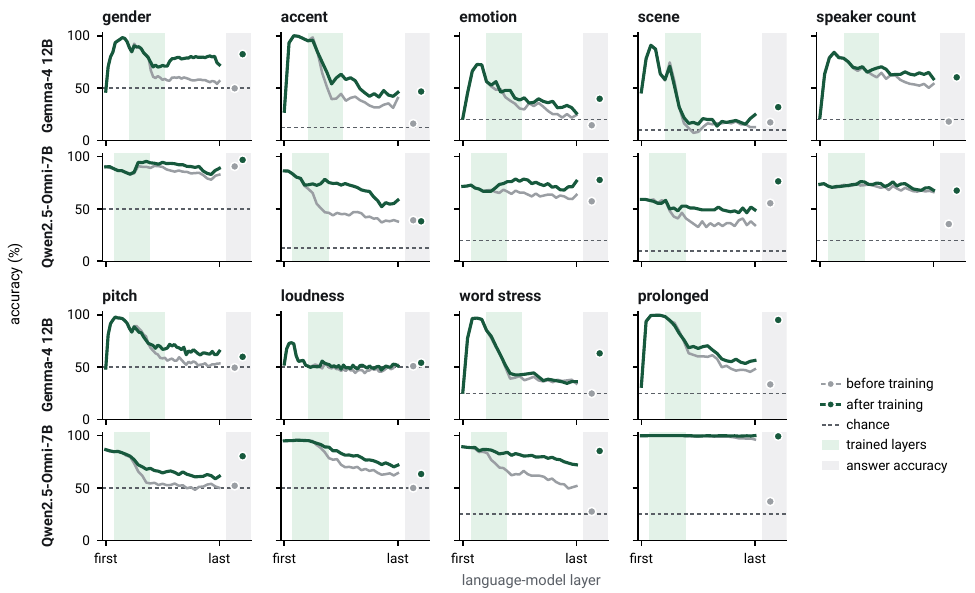}
% 中文：训练前后的逐层探针准确率与作答准确率。上排为 Gemma-4 12B、下排为 Qwen2.5-Omni-7B，灰线和绿线分别表示训练前后，其余标记同图~\ref{fig:probe}。口音面板为 AccentDB；第~\ref{sec:repair} 节平均的 11 个探针任务还包括未画出的 VoxCeleb 和 MMSU 口音任务。
\caption{Layer-wise probe and answer accuracy before and after training. Upper panels show Gemma-4 12B and lower panels Qwen2.5-Omni-7B, with grey and green lines for probe accuracy before and after training and other markers as in Figure~\ref{fig:probe}. The accent panel shows AccentDB, and the 11 probed tasks averaged in Section~\ref{sec:repair} also include the VoxCeleb and MMSU accent tasks, which are not drawn.}
\label{fig:repair-curves}
\end{figure*}

% 中文：图~\ref{fig:probe-all} 展示两个模型在所有探针任务上的逐层准确率。图~\ref{fig:repair-curves} 则比较训练前后的探针读出与作答准确率。
Figure~\ref{fig:probe-all} shows probe accuracy across layers for both models and all probed tasks. Figure~\ref{fig:repair-curves} compares probe readout and answer accuracy before and after training.

%% file: sec/app_g_training.tex
% 中文：训练数据与设置
\section{Training Data and Setup}
\label{app:g}

% 中文：训练集包含 219.0 小时音频，约 80\% 来自 LibriSpeech 的 train-clean-100 和 train-clean-360。其余音频来自 16 个语料库。九个语音语料库包括 Expresso \citep{nguyen2023expresso}、Speech Accent Archive \citep{weinberger2015accent}、EARS \citep{richter2024ears}、speechocean762 \citep{zhang2021speechocean}、AfriSpeech-200 \citep{olatunji2023afrispeech}、NISP \citep{kalluri2021nisp}、ARU \citep{hopkins2019aru}、JIBO Kids \citep{shankar2024jibo} 和儿童录音 \citep{kennedy2016children}。四个表演情感语料库是 CaFE \citep{gournay2018cafe}、TESS \citep{pichorafuller2020tess}、EmoV-DB \citep{adigwe2018emovdb} 和 JL-Corpus \citep{james2018jlcorpus}。三个声音事件语料库是 FSD50K \citep{fonseca2022fsd50k}、Nonspeech7k \citep{rashid2023nonspeech7k} 和 ESC-50 \citep{piczak2015esc}。
The training data contain 219.0 hours of audio, roughly 80\% from the LibriSpeech train-clean-100 and train-clean-360 splits. The remaining audio comes from 16 corpora. The nine speech corpora include Expresso \citep{nguyen2023expresso}, Speech Accent Archive \citep{weinberger2015accent}, EARS \citep{richter2024ears}, speechocean762 \citep{zhang2021speechocean}, AfriSpeech-200 \citep{olatunji2023afrispeech}, NISP \citep{kalluri2021nisp}, ARU \citep{hopkins2019aru}, JIBO Kids \citep{shankar2024jibo}, and children's recordings \citep{kennedy2016children}. The four acted-emotion corpora are CaFE \citep{gournay2018cafe}, TESS \citep{pichorafuller2020tess}, EmoV-DB \citep{adigwe2018emovdb}, and JL-Corpus \citep{james2018jlcorpus}. The three sound-event corpora are FSD50K \citep{fonseca2022fsd50k}, Nonspeech7k \citep{rashid2023nonspeech7k}, and ESC-50 \citep{piczak2015esc}.

% 中文：一部分训练题由改动录音生成。我们调整 LibriSpeech 和 Expresso 语句的音高、时长、噪声或混响，或将不同说话人的片段拼成对话。这些题的答案由改动方式确定。我们用 pYIN \citep{mauch2014pyin} 估计音高，用 Montreal Forced Aligner \citep{mcauliffe2017mfa} 定位词边界，用 Pyroomacoustics \citep{scheibler2018pyroomacoustics} 模拟房间混响。其他题的答案来自语料元数据、表演情感标注或声音事件标注。
We generated some training items by changing the recordings. We adjusted the pitch, duration, noise, or reverberation of LibriSpeech and Expresso utterances, or spliced segments from different speakers into dialogues. Each change determines the answer. We estimated pitch with pYIN \citep{mauch2014pyin}, located word boundaries with Montreal Forced Aligner \citep{mcauliffe2017mfa}, and simulated room reverberation with Pyroomacoustics \citep{scheibler2018pyroomacoustics}. Other answers come from corpus metadata, acted-emotion labels, or sound-event labels.

% 中文：训练题覆盖 34 个评测任务的题型，其中 23 个是属性类、11 个是结构类。另有 8 个属性与结构任务没有同题型训练题，分别是 7 个对话情感与讽刺任务，以及 MMSU 响度比较。表~\ref{tab:h1} 将它们标为灰色。词汇类和混合类任务也没有同题型训练题。
Training items use the question formats of 34 evaluation tasks, comprising 23 attribute and 11 structure tasks. Eight other attribute and structure tasks have no training items in the same format. These are seven conversational emotion or sarcasm tasks and MMSU loudness comparison, shaded grey in Table~\ref{tab:h1}. No lexical or mixed task has training items in the same format either. 

% 中文：对于这 34 种题型，训练题沿用评测的提示模板、选项数和标签词。约一半直接使用对应评测任务的通用问句。部分重音、延长音和词语调题改写了问法，部分情绪题列出了候选标签。训练中没有使用评测题的录音或题目特有的文字，也没有使用评测语料的录音。我们还剔除了音频指纹或句子与评测题重合的录音。训练和评测因此只共享题型。
For these 34 formats, training items follow the evaluation prompt templates, option counts, and label words. About half reuse the evaluation task's generic question. Some stress, prolonged-sound, and word-intonation items rephrase the question, while some emotion items list the candidate labels. We used no evaluation audio, item-specific evaluation text, or recordings from the evaluation corpora. We also removed training recordings whose audio fingerprint or sentence matched an evaluation item. Training and evaluation therefore share question formats, but not items.

% 中文：我们在读出下降层的所有注意力和前馈投影上训练秩为 64 的 LoRA 适配器。AdamW 训练共进行 17,548 步，每步 8 道题，最高学习率为 $3\times10^{-5}$。结果均使用最后一步的适配器，没有依据评测准确率选择检查点。
We trained rank-64 LoRA adapters on every attention and feed-forward projection in the readout-decline layers. We used AdamW for 17{,}548 steps with 8 items per step and a peak learning rate of $3\times10^{-5}$. All results use the final adapter, with no checkpoint selection based on evaluation accuracy.

%% file: sec/app_h_repair.tex
\onecolumn % longtable cannot run in two-column mode; this appendix is set in one column
% 中文：逐任务修复结果
\section{Per-Task Repair Results}
\label{app:h}

\begin{figure}[t]
\centering
\includegraphics[width=\linewidth]{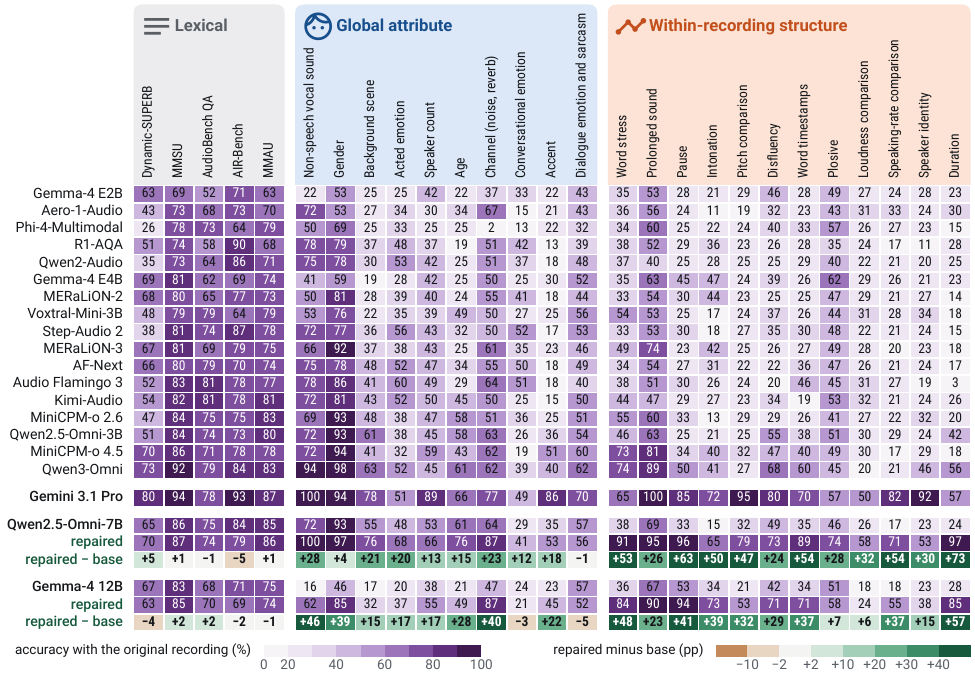}
% 中文：各模型在原录音上的分组准确率。图中包括 19 个开源模型、Gemini 3.1 Pro 和两个修复后模型，紫色表示准确率（%），绿色和棕色表示修复后相对基础模型的变化（百分点），词汇类按基准分组。
\caption{Accuracy with the original recording by task group. Rows cover 19 open models, Gemini 3.1 Pro, and two repaired models, with purple cells for accuracy (\%), green and brown cells for repaired-minus-base changes (pp), and lexical tasks grouped by benchmark.}
\label{fig:gap-matrix}
\end{figure}

% 中文：表~\ref{tab:h1} 列出两个模型训练前后在全部 92 个任务上的原录音准确率。图~\ref{fig:gap-matrix} 将这些任务按组汇总。灰色行标出 8 个没有同题型训练题的属性与结构任务，可用于观察训练效果能否迁移到这些题型。
Table~\ref{tab:h1} reports accuracy with the original recording before and after training for both models on all 92 tasks. Figure~\ref{fig:gap-matrix} groups these tasks. The eight shaded attribute and structure tasks have no training items in the same format, allowing us to examine transfer to these formats.

{\footnotesize
\setlength{\tabcolsep}{4pt}
\setlength{\LTcapwidth}{\linewidth}
\makeatletter % caption at the size of the other captions, table body in footnotesize
\let\LTcapOrig\LT@makecaption
\def\LT@makecaption#1#2#3{\LTcapOrig{#1}{\normalsize #2}{\normalsize #3}}
\makeatother
\begin{longtable}{l>{\raggedright\arraybackslash}p{0.45\linewidth}rrrr}
% 中文：两个模型训练前后在 92 个任务上的原录音准确率（%）。任务按表~\ref{tab:b2} 分类，修复后单元格按相对基础模型的变化着色（深绿表示较大提升，棕色表示下降，浅灰表示变化不超过 2 个百分点），灰色行标出没有同题型训练题的属性与结构任务。
\caption{Original-recording accuracy (\%) before and after training on 92 tasks. Tasks follow the classes in Table~\ref{tab:b2}, with repaired cells colored by change from the base (darker green for larger gains, brown for declines, light grey within 2\,pp) and shaded rows marking attribute and structure tasks without matching training formats.}\label{tab:h1}\\
\toprule
 & & \multicolumn{2}{c}{Gemma-4 12B} & \multicolumn{2}{c}{Qwen2.5-Omni-7B} \\
\cmidrule(lr){3-4}\cmidrule(lr){5-6}
Benchmark & Task & \multicolumn{1}{c}{Base} & \multicolumn{1}{c}{Repaired} & \multicolumn{1}{c}{Base} & \multicolumn{1}{c}{Repaired} \\
\midrule
\endfirsthead
\toprule
 & & \multicolumn{2}{c}{Gemma-4 12B} & \multicolumn{2}{c}{Qwen2.5-Omni-7B} \\
\cmidrule(lr){3-4}\cmidrule(lr){5-6}
Benchmark & Task & \multicolumn{1}{c}{Base} & \multicolumn{1}{c}{Repaired} & \multicolumn{1}{c}{Base} & \multicolumn{1}{c}{Repaired} \\
\midrule
\endhead
\bottomrule
\endfoot
\multicolumn{6}{l}{\emph{Lexical (36 tasks)}} \\
AIR-Bench & Speaker\_Intent\_Classification & 73.9 & \cellcolor[HTML]{E8D6C2}70.4 & 90.8 & \cellcolor[HTML]{E8D6C2}85.2 \\
AIR-Bench & Speech\_Entity\_Reconition & 67.1 & \cellcolor[HTML]{F4F4F2}67.8 & 77.6 & \cellcolor[HTML]{E8D6C2}73.4 \\
AudioBench & cn\_college\_listen\_mcq\_test & 89.6 & \cellcolor[HTML]{F4F4F2}91.4 & 92.6 & \cellcolor[HTML]{F4F4F2}92.6 \\
AudioBench & dream\_tts\_mcq\_test & 77.0 & \cellcolor[HTML]{F4F4F2}78.2 & 89.9 & \cellcolor[HTML]{F4F4F2}89.1 \\
AudioBench & public\_sg\_speech\_qa\_test & 22.3 & \cellcolor[HTML]{D3E6DA}26.6 & 33.2 & \cellcolor[HTML]{F4F4F2}32.2 \\
AudioBench & slue\_p2\_sqa5\_test & 81.7 & \cellcolor[HTML]{F4F4F2}82.9 & 83.2 & \cellcolor[HTML]{F4F4F2}81.2 \\
Dynamic-SUPERB & DialogueActClassification\_DailyTalk & 62.7 & \cellcolor[HTML]{E8D6C2}57.5 & 50.3 & \cellcolor[HTML]{A3CFB8}64.2 \\
Dynamic-SUPERB & SuperbIC\_SLURP-Intent & 71.2 & \cellcolor[HTML]{E8D6C2}69.1 & 80.1 & \cellcolor[HTML]{E8D6C2}74.9 \\
MMAU & Conversational Fact Retrieval & 80.0 & \cellcolor[HTML]{F4F4F2}80.0 & 100.0 & \cellcolor[HTML]{E8D6C2}95.0 \\
MMAU & Dissonant Emotion Interpretation & 95.0 & \cellcolor[HTML]{E8D6C2}90.0 & 100.0 & \cellcolor[HTML]{F4F4F2}100.0 \\
MMAU & Event-Based Knowledge Retrieval & 75.0 & \cellcolor[HTML]{C58A5A}60.0 & 90.0 & \cellcolor[HTML]{E8D6C2}85.0 \\
MMAU & Key highlight Extraction & 95.0 & \cellcolor[HTML]{E8D6C2}85.0 & 95.0 & \cellcolor[HTML]{F4F4F2}95.0 \\
MMAU & Multi Speaker Role Mapping & 100.0 & \cellcolor[HTML]{F4F4F2}100.0 & 100.0 & \cellcolor[HTML]{F4F4F2}100.0 \\
MMAU & Phonemic Stress Pattern Analysis & 34.6 & \cellcolor[HTML]{A3CFB8}45.1 & 54.9 & \cellcolor[HTML]{D3E6DA}57.5 \\
MMAU & Phonological Sequence Decoding & 47.4 & \cellcolor[HTML]{A3CFB8}57.9 & 52.6 & \cellcolor[HTML]{A3CFB8}68.4 \\
MMSU & casual\_reasoning & 98.3 & \cellcolor[HTML]{F4F4F2}100.0 & 100.0 & \cellcolor[HTML]{F4F4F2}100.0 \\
MMSU & code\_switch\_question\_answering & 67.5 & \cellcolor[HTML]{A3CFB8}79.2 & 94.8 & \cellcolor[HTML]{F4F4F2}93.5 \\
MMSU & consonant\_vowel\_perception & 75.4 & \cellcolor[HTML]{D3E6DA}78.7 & 60.7 & \cellcolor[HTML]{D3E6DA}70.5 \\
MMSU & content\_grounding & 95.4 & \cellcolor[HTML]{D3E6DA}98.2 & 93.5 & \cellcolor[HTML]{F4F4F2}94.4 \\
MMSU & continuation\_writing & 98.2 & \cellcolor[HTML]{F4F4F2}100.0 & 99.1 & \cellcolor[HTML]{F4F4F2}100.0 \\
MMSU & couplet\_matching & 40.6 & \cellcolor[HTML]{D3E6DA}46.5 & 66.3 & \cellcolor[HTML]{E8D6C2}59.4 \\
MMSU & deixis\_resolution & 84.5 & \cellcolor[HTML]{F4F4F2}82.5 & 90.3 & \cellcolor[HTML]{F4F4F2}91.3 \\
MMSU & dialogue\_reasoning & 82.3 & \cellcolor[HTML]{D3E6DA}86.7 & 87.6 & \cellcolor[HTML]{F4F4F2}85.8 \\
MMSU & homophone\_based\_reasoning & 93.1 & \cellcolor[HTML]{E8D6C2}89.1 & 98.0 & \cellcolor[HTML]{F4F4F2}96.0 \\
MMSU & idiom\_reasoning & 96.0 & \cellcolor[HTML]{F4F4F2}97.0 & 100.0 & \cellcolor[HTML]{F4F4F2}100.0 \\
MMSU & intent\_detection & 78.5 & \cellcolor[HTML]{D3E6DA}85.0 & 89.7 & \cellcolor[HTML]{E8D6C2}83.2 \\
MMSU & logical\_reasoning & 73.1 & \cellcolor[HTML]{F4F4F2}74.0 & 78.8 & \cellcolor[HTML]{F4F4F2}77.9 \\
MMSU & long\_speech\_summarization & 97.2 & \cellcolor[HTML]{F4F4F2}99.1 & 99.1 & \cellcolor[HTML]{F4F4F2}99.1 \\
MMSU & near\_homophone\_perception & 52.8 & \cellcolor[HTML]{D3E6DA}55.6 & 68.1 & \cellcolor[HTML]{F4F4F2}68.1 \\
MMSU & polysemy\_reasoning & 96.9 & \cellcolor[HTML]{F4F4F2}95.9 & 99.0 & \cellcolor[HTML]{F4F4F2}99.0 \\
MMSU & prolonged\_sound\_reasoning & 93.9 & \cellcolor[HTML]{F4F4F2}93.9 & 97.6 & \cellcolor[HTML]{F4F4F2}98.8 \\
MMSU & puns\_interpretation & 100.0 & \cellcolor[HTML]{F4F4F2}100.0 & 99.1 & \cellcolor[HTML]{F4F4F2}100.0 \\
MMSU & speech\_act\_classification & 94.8 & \cellcolor[HTML]{D3E6DA}97.9 & 90.7 & \cellcolor[HTML]{D3E6DA}92.8 \\
MMSU & speech\_translation & 94.0 & \cellcolor[HTML]{F4F4F2}92.8 & 97.6 & \cellcolor[HTML]{F4F4F2}97.6 \\
MMSU & syllable\_perception & 38.1 & \cellcolor[HTML]{E8D6C2}35.7 & 27.4 & \cellcolor[HTML]{A3CFB8}44.0 \\
MMSU & syntactic\_structure\_matching & 94.8 & \cellcolor[HTML]{D3E6DA}97.9 & 73.2 & \cellcolor[HTML]{E8D6C2}68.0 \\
\midrule
\multicolumn{6}{l}{\emph{Global attribute (30 tasks)}} \\
AIR-Bench & Speaker\_Age\_Prediction & 19.1 & \cellcolor[HTML]{3C8C67}\textcolor{white}{55.3} & 46.8 & \cellcolor[HTML]{6BB08D}74.5 \\
AIR-Bench & Speaker\_Emotion\_Recontion & 44.1 & \cellcolor[HTML]{F4F4F2}43.4 & 57.3 & \cellcolor[HTML]{D3E6DA}67.1 \\
AIR-Bench & Speaker\_Gender\_Recognition & 41.0 & \cellcolor[HTML]{17593D}\textcolor{white}{84.0} & 93.8 & \cellcolor[HTML]{D3E6DA}95.8 \\
AIR-Bench & Speaker\_Number\_Verification & 33.3 & \cellcolor[HTML]{3C8C67}\textcolor{white}{71.3} & 64.0 & \cellcolor[HTML]{A3CFB8}78.0 \\
AudioBench & iemocap\_emotion\_recognition & 11.8 & \cellcolor[HTML]{6BB08D}38.9 & 47.8 & \cellcolor[HTML]{D3E6DA}57.6 \\
AudioBench & iemocap\_gender\_recognition & 52.4 & \cellcolor[HTML]{6BB08D}75.5 & 96.5 & \cellcolor[HTML]{E8D6C2}93.4 \\
\rowcolor{black!8}
AudioBench & meld\_emotion\_test & 10.9 & \cellcolor[HTML]{D3E6DA}14.5 & 23.7 & \cellcolor[HTML]{A3CFB8}38.0 \\
\rowcolor{black!8}
AudioBench & meld\_sentiment\_test & 19.2 & \cellcolor[HTML]{F4F4F2}20.9 & 12.2 & \cellcolor[HTML]{A3CFB8}27.2 \\
AudioBench & voxceleb\_accent\_test & 32.9 & \cellcolor[HTML]{3C8C67}\textcolor{white}{65.0} & 55.0 & \cellcolor[HTML]{6BB08D}75.9 \\
AudioBench & voxceleb\_gender\_test & 47.5 & \cellcolor[HTML]{17593D}\textcolor{white}{89.8} & 85.3 & \cellcolor[HTML]{A3CFB8}100.0 \\
Dynamic-SUPERB & AccentClassification\_AccentdbExtended & 16.1 & \cellcolor[HTML]{3C8C67}\textcolor{white}{46.7} & 39.2 & \cellcolor[HTML]{F4F4F2}38.2 \\
\rowcolor{black!8}
Dynamic-SUPERB & DialogueEmotionClassification\_DailyTalk & 63.3 & \cellcolor[HTML]{C58A5A}51.0 & 72.5 & \cellcolor[HTML]{D3E6DA}75.0 \\
\rowcolor{black!8}
Dynamic-SUPERB & EmotionRecognition\_MultimodalEmotionlinesDataset & 40.6 & \cellcolor[HTML]{C58A5A}27.6 & 52.1 & \cellcolor[HTML]{D3E6DA}57.8 \\
Dynamic-SUPERB & MultiSpeakerDetection\_VCTK & 51.5 & \cellcolor[HTML]{17593D}\textcolor{white}{99.0} & 78.5 & \cellcolor[HTML]{6BB08D}99.0 \\
Dynamic-SUPERB & NoiseDetection\_VCTK\_MUSAN-Noise & 47.2 & \cellcolor[HTML]{3C8C67}\textcolor{white}{84.6} & 67.2 & \cellcolor[HTML]{A3CFB8}83.6 \\
Dynamic-SUPERB & ReverberationDetection\_VCTK\_RirsNoises-MediumRoom & 47.7 & \cellcolor[HTML]{17593D}\textcolor{white}{89.3} & 61.4 & \cellcolor[HTML]{6BB08D}89.8 \\
\rowcolor{black!8}
Dynamic-SUPERB & SarcasmDetection\_Mustard & 62.0 & \cellcolor[HTML]{D3E6DA}67.0 & 62.0 & \cellcolor[HTML]{F4F4F2}62.0 \\
Dynamic-SUPERB & SpeakerCounting\_LibriTTS-TestClean & 18.0 & \cellcolor[HTML]{17593D}\textcolor{white}{60.3} & 35.6 & \cellcolor[HTML]{3C8C67}\textcolor{white}{67.5} \\
MMAU & Counting & 50.0 & \cellcolor[HTML]{C58A5A}35.0 & 55.0 & \cellcolor[HTML]{A3CFB8}70.0 \\
\rowcolor{black!8}
MMAU & Emotion Flip Detection & 50.0 & \cellcolor[HTML]{C58A5A}30.0 & 30.0 & \cellcolor[HTML]{E8D6C2}25.0 \\
MMSU & accent\_identification & 29.2 & \cellcolor[HTML]{A3CFB8}42.5 & 30.1 & \cellcolor[HTML]{3C8C67}\textcolor{white}{68.1} \\
MMSU & age\_prediction & 22.6 & \cellcolor[HTML]{A3CFB8}42.2 & 75.5 & \cellcolor[HTML]{D3E6DA}77.5 \\
MMSU & background\_scene\_recognition & 17.3 & \cellcolor[HTML]{A3CFB8}31.8 & 55.5 & \cellcolor[HTML]{6BB08D}76.4 \\
MMSU & dialogue\_turn\_counting & 40.4 & \cellcolor[HTML]{C58A5A}24.2 & 50.5 & \cellcolor[HTML]{C58A5A}35.4 \\
MMSU & emotion\_recognition & 14.6 & \cellcolor[HTML]{6BB08D}39.8 & 57.3 & \cellcolor[HTML]{6BB08D}77.7 \\
MMSU & emotional\_context\_reasoning & 10.0 & \cellcolor[HTML]{A3CFB8}27.8 & 28.9 & \cellcolor[HTML]{17593D}\textcolor{white}{68.9} \\
MMSU & gender\_prediction & 44.9 & \cellcolor[HTML]{17593D}\textcolor{white}{89.0} & 97.5 & \cellcolor[HTML]{F4F4F2}97.5 \\
MMSU & non\_verbal\_sound\_detection & 15.6 & \cellcolor[HTML]{17593D}\textcolor{white}{62.5} & 71.9 & \cellcolor[HTML]{6BB08D}100.0 \\
\rowcolor{black!8}
MMSU & sarcasm\_detection & 52.2 & \cellcolor[HTML]{D3E6DA}58.6 & 64.0 & \cellcolor[HTML]{E8D6C2}61.3 \\
MMSU & total\_speaker\_counting & 34.8 & \cellcolor[HTML]{D3E6DA}40.7 & 33.9 & \cellcolor[HTML]{A3CFB8}46.6 \\
\midrule
\multicolumn{6}{l}{\emph{Within-recording structure (12 tasks)}} \\
AIR-Bench & Speech\_Grounding & 34.2 & \cellcolor[HTML]{3C8C67}\textcolor{white}{71.1} & 34.9 & \cellcolor[HTML]{17593D}\textcolor{white}{88.6} \\
MMSU & disfluency\_detection & 42.5 & \cellcolor[HTML]{6BB08D}70.8 & 49.1 & \cellcolor[HTML]{6BB08D}72.6 \\
MMSU & intonation\_perception & 33.6 & \cellcolor[HTML]{3C8C67}\textcolor{white}{72.9} & 14.9 & \cellcolor[HTML]{17593D}\textcolor{white}{65.4} \\
MMSU & pause\_perception & 52.8 & \cellcolor[HTML]{17593D}\textcolor{white}{94.3} & 33.0 & \cellcolor[HTML]{17593D}\textcolor{white}{96.2} \\
MMSU & pitch\_comparison & 21.2 & \cellcolor[HTML]{3C8C67}\textcolor{white}{52.5} & 32.3 & \cellcolor[HTML]{17593D}\textcolor{white}{78.8} \\
MMSU & plosive\_sound\_identification & 50.5 & \cellcolor[HTML]{D3E6DA}57.7 & 46.4 & \cellcolor[HTML]{6BB08D}74.2 \\
MMSU & prolonged\_sound\_perception & 66.7 & \cellcolor[HTML]{6BB08D}90.1 & 69.1 & \cellcolor[HTML]{6BB08D}95.1 \\
MMSU & speaker\_identity\_recognition & 22.8 & \cellcolor[HTML]{A3CFB8}37.7 & 22.8 & \cellcolor[HTML]{6BB08D}52.6 \\
MMSU & speech\_duration\_estimation & 27.6 & \cellcolor[HTML]{17593D}\textcolor{white}{84.8} & 23.8 & \cellcolor[HTML]{17593D}\textcolor{white}{97.1} \\
MMSU & speech\_stress\_perception & 36.4 & \cellcolor[HTML]{17593D}\textcolor{white}{83.6} & 38.2 & \cellcolor[HTML]{17593D}\textcolor{white}{90.9} \\
MMSU & speed\_comparison & 18.5 & \cellcolor[HTML]{3C8C67}\textcolor{white}{55.4} & 17.4 & \cellcolor[HTML]{17593D}\textcolor{white}{70.7} \\
\rowcolor{black!8}
MMSU & volume\_comparison & 17.8 & \cellcolor[HTML]{D3E6DA}23.8 & 25.7 & \cellcolor[HTML]{3C8C67}\textcolor{white}{58.4} \\
\midrule
\multicolumn{6}{l}{\emph{Mixed (14 tasks)}} \\
AIR-Bench & Synthesized\_Voice\_Detection & 48.3 & \cellcolor[HTML]{D3E6DA}55.0 & 55.0 & \cellcolor[HTML]{F4F4F2}55.7 \\
Dynamic-SUPERB & SpoofDetection\_ASVspoof2017 & 7.9 & \cellcolor[HTML]{17593D}\textcolor{white}{59.5} & 10.1 & \cellcolor[HTML]{D3E6DA}18.5 \\
MMAU & Emotion State summarisation & 80.0 & \cellcolor[HTML]{C58A5A}65.0 & 25.0 & \cellcolor[HTML]{C58A5A}10.0 \\
MMAU-Pro & Audio Quality, Artifacts \& Channel Characteristics & 60.0 & \cellcolor[HTML]{C58A5A}40.0 & 80.0 & \cellcolor[HTML]{F4F4F2}80.0 \\
MMAU-Pro & Lexical and Phrase-Level Recognition & 100.0 & \cellcolor[HTML]{F4F4F2}100.0 & 100.0 & \cellcolor[HTML]{F4F4F2}100.0 \\
MMAU-Pro & Paralinguistic/Emotion Recognition & 67.2 & \cellcolor[HTML]{D3E6DA}70.5 & 72.1 & \cellcolor[HTML]{E8D6C2}67.2 \\
MMAU-Pro & Speaker Demographics & 48.1 & \cellcolor[HTML]{F4F4F2}50.0 & 62.7 & \cellcolor[HTML]{F4F4F2}62.0 \\
MMAU-Pro & Speaker Identification & 75.0 & \cellcolor[HTML]{A3CFB8}87.5 & 100.0 & \cellcolor[HTML]{F4F4F2}100.0 \\
MMAU-Pro & Speech Activity, Turn-Taking and Overlap Detection & 56.1 & \cellcolor[HTML]{E8D6C2}53.7 & 63.4 & \cellcolor[HTML]{D3E6DA}65.8 \\
MMSU & intonation\_based\_reasoning & 63.6 & \cellcolor[HTML]{6BB08D}85.7 & 74.0 & \cellcolor[HTML]{A3CFB8}87.0 \\
MMSU & pause\_based\_reasoning & 50.6 & \cellcolor[HTML]{D3E6DA}55.1 & 53.9 & \cellcolor[HTML]{F4F4F2}52.8 \\
MMSU & stress\_based\_reasoning & 50.0 & \cellcolor[HTML]{6BB08D}72.7 & 47.3 & \cellcolor[HTML]{A3CFB8}64.5 \\
MMSU & synthetic\_speech\_detection & 43.1 & \cellcolor[HTML]{F4F4F2}43.1 & 56.0 & \cellcolor[HTML]{D3E6DA}58.7 \\
MMSU & vocal\_range\_comparison & 50.0 & \cellcolor[HTML]{C58A5A}37.5 & 62.5 & \cellcolor[HTML]{F4F4F2}62.5 \\
\end{longtable}
}
\twocolumn

%% file: refs.bib
@inproceedings{sakshi2024mmau,
  title={{MMAU}: A Massive Multi-Task Audio Understanding and Reasoning Benchmark},
  author={Sakshi, S and Tyagi, Utkarsh and Kumar, Sonal and Seth, Ashish and Selvakumar, Ramaneswaran and Nieto, Oriol and Duraiswami, Ramani and Ghosh, Sreyan and Manocha, Dinesh},
  booktitle={International Conference on Learning Representations},
  year={2025},
  note={arXiv:2410.19168}
}

@inproceedings{wang2026mmsu,
  title={{MMSU}: A Massive Multi-task Spoken Language Understanding and Reasoning Benchmark},
  author={Wang, Dingdong and Li, Junan and Wu, Jincenzi and Yang, Dongchao and Chen, Xueyuan and Zhang, Tianhua and Meng, Helen},
  booktitle={International Conference on Learning Representations},
  year={2026}
}

@inproceedings{kumar2025mmaupro,
  title={{MMAU-Pro}: A Challenging and Comprehensive Benchmark for Holistic Evaluation of Audio General Intelligence},
  author={Kumar, Sonal and Sedl{\'a}{\v{c}}ek, {\v{S}}imon and Lokegaonkar, Vaibhavi and L{\'o}pez, Fernando and Yu, Wenyi and Anand, Nishit and others},
  booktitle={AAAI Conference on Artificial Intelligence},
  volume={40},
  pages={22688--22697},
  year={2026},
  doi={10.1609/aaai.v40i27.39430}
}

@inproceedings{wang2024audiobench,
  title={{AudioBench}: A Universal Benchmark for Audio Large Language Models},
  author={Wang, Bin and Zou, Xunlong and Lin, Geyu and Sun, Shuo and Liu, Zhuohan and Zhang, Wenyu and Liu, Zhengyuan and Aw, AiTi and Chen, Nancy F.},
  booktitle={Conference of the Nations of the Americas Chapter of the Association for Computational Linguistics},
  pages={4297--4316},
  year={2025},
  doi={10.18653/v1/2025.naacl-long.218}
}

@inproceedings{yang2024airbench,
  title={{AIR-Bench}: Benchmarking Large Audio-Language Models via Generative Comprehension},
  author={Yang, Qian and Xu, Jin and Liu, Wenrui and Chu, Yunfei and Jiang, Ziyue and Zhou, Xiaohuan and Leng, Yichong and Lv, Yuanjun and Zhao, Zhou and Zhou, Chang and Zhou, Jingren},
  booktitle={Annual Meeting of the Association for Computational Linguistics},
  year={2024}
}

@inproceedings{huang2025dynamicsuperb,
  title={{Dynamic-SUPERB} Phase-2: A Collaboratively Expanding Benchmark for Measuring the Capabilities of Spoken Language Models with 180 Tasks},
  author={Huang, Chien-yu and Chen, Wei-Chih and Yang, Shu-wen and Liu, Andy T. and Li, Chen-An and Lin, Yu-Xiang and others},
  booktitle={International Conference on Learning Representations},
  year={2025},
  note={arXiv:2411.05361}
}

@article{foo2026glitters,
  title={All That Glitters Is Not Audio: Rethinking Text Priors and Audio Reliance in Audio-Language Evaluation},
  author={Foo, Leonardo Haw-Yang and Yang, Chih-Kai and Li, Chen-An and Lu, Ke-Han and Lee, Hung-yi},
  journal={arXiv preprint arXiv:2604.24401},
  year={2026}
}

@inproceedings{chen2025listen,
  title={Do Audio {LLMs} Really {LISTEN}, or Just Transcribe? {Measuring} Lexical vs. Acoustic Emotion Cues Reliance},
  author={Chen, Jingyi and Guo, Zhimeng and Chun, Jiyun and Wang, Pichao and Perrault, Andrew and Elsner, Micha},
  booktitle={Conference of the European Chapter of the Association for Computational Linguistics},
  year={2026},
  note={arXiv:2510.10444}
}

@article{xiong2026deaf,
  title={{DEAF}: A Benchmark for Diagnostic Evaluation of Acoustic Faithfulness in Audio Language Models},
  author={Xiong, Jiaqi and Qi, Yunjia and Cao, Qi and Zheng, Yu and Zhang, Yutong and Wang, Ziteng and Liao, Ruofan and Xu, Weisheng and Liu, Sichen},
  journal={arXiv preprint arXiv:2603.18048},
  year={2026}
}

@inproceedings{wang2025mcr,
  title={When Audio and Text Disagree: Revealing Text Bias in Large Audio-Language Models},
  author={Wang, Cheng and Deng, Gelei and Yang, Xianglin and Qiu, Han and Zhang, Tianwei},
  booktitle={Conference on Empirical Methods in Natural Language Processing},
  year={2025},
  note={arXiv:2508.15407}
}

@inproceedings{pang2026voxparadox,
  title={Do Audio {LLMs} Listen or Read? {Analyzing} and Mitigating Paralinguistic Failures with {VoxParadox}},
  author={Pang, Jiacheng and Chaubey, Ashutosh and Soleymani, Mohammad},
  booktitle={International Conference on Machine Learning},
  year={2026},
  note={arXiv:2605.27772}
}

@article{billa2026alme,
  title={When Audio-{LLMs} Don't Listen: A Cross-Linguistic Study of Modality Arbitration},
  author={Billa, Jayadev},
  journal={arXiv preprint arXiv:2602.11488},
  year={2026}
}

@article{sun2026sonicbench,
  title={{SonicBench}: Dissecting the Physical Perception Bottleneck in Large Audio Language Models},
  author={Sun, Yirong and Chen, Yanjun and Qiu, Xin and Zhang, Gang and Chen, Hongyu and Wu, Daokuan and Li, Chengming and Yang, Min and Zhu, Dawei and Zhang, Wei and Shen, Xiaoyu},
  journal={arXiv preprint arXiv:2601.11039},
  year={2026}
}

@article{gao2026arbitration,
  title={Beyond Text Following: Repairable Arbitration Reversals in Audio-Language Models},
  author={Gao, Yichen and Zhang, Yiqun and Wang, Zijing and Li, Yujia and Guo, Heng and Wu, Xi and Yang, Xiaocui and Feng, Shi and Zhang, Yifei and Wang, Daling},
  journal={arXiv preprint arXiv:2606.05161},
  year={2026}
}

@inproceedings{radford2022whisper,
  title={Robust Speech Recognition via Large-Scale Weak Supervision},
  author={Radford, Alec and Kim, Jong Wook and Xu, Tao and Brockman, Greg and McLeavey, Christine and Sutskever, Ilya},
  booktitle={International Conference on Machine Learning},
  pages={28492--28518},
  year={2023},
  note={arXiv:2212.04356}
}

@misc{edgetts,
  title={edge-tts: Use {Microsoft Edge}'s Online Text-to-Speech Service from {Python}},
  author={{rany2}},
  year={2024},
  howpublished={\url{https://github.com/rany2/edge-tts}}
}

@article{chu2024qwen2audio,
  title={{Qwen2-Audio} Technical Report},
  author={Chu, Yunfei and Xu, Jin and Yang, Qian and Wei, Haojie and Wei, Xipin and Guo, Zhifang and others},
  journal={arXiv preprint arXiv:2407.10759},
  year={2024}
}

@article{xu2025qwen25omni,
  title={{Qwen2.5-Omni} Technical Report},
  author={Xu, Jin and Guo, Zhifang and He, Jinzheng and Hu, Hangrui and He, Ting and Bai, Shuai and others},
  journal={arXiv preprint arXiv:2503.20215},
  year={2025}
}

@article{xu2025qwen3omni,
  title={{Qwen3-Omni} Technical Report},
  author={Xu, Jin and Guo, Zhifang and Hu, Hangrui and Chu, Yunfei and Wang, Xiong and He, Jinzheng and others},
  journal={arXiv preprint arXiv:2509.17765},
  year={2025}
}

@article{kimiteam2025kimiaudio,
  title={{Kimi-Audio} Technical Report},
  author={{KimiTeam} and Ding, Ding and Ju, Zeqian and Leng, Yichong and Liu, Songxiang and Liu, Tong and others},
  journal={arXiv preprint arXiv:2504.18425},
  year={2025}
}

@article{abouelenin2025phi4mini,
  title={{Phi-4-Mini} Technical Report: Compact yet Powerful Multimodal Language Models via Mixture-of-{LoRAs}},
  author={Abouelenin, Abdelrahman and Ashfaq, Atabak and Atkinson, Adam and Awadalla, Hany and Bach, Nguyen and others},
  journal={arXiv preprint arXiv:2503.01743},
  year={2025}
}

@article{yao2024minicpmv,
  title={{MiniCPM-V}: A {GPT-4V} Level {MLLM} on Your Phone},
  author={Yao, Yuan and Yu, Tianyu and Zhang, Ao and Wang, Chongyi and Cui, Junbo and Zhu, Hongji and others},
  journal={arXiv preprint arXiv:2408.01800},
  year={2024}
}

@inproceedings{goel2025af3,
  title={{Audio Flamingo 3}: Advancing Audio Intelligence with Fully Open Large Audio Language Models},
  author={Ghosh, Sreyan and Goel, Arushi and Kim, Jaehyeon and Kumar, Sonal and Kong, Zhifeng and Lee, Sang-gil and others},
  booktitle={Advances in Neural Information Processing Systems},
  year={2025},
  note={arXiv:2507.08128}
}

@article{li2025r1aqa,
  title={Reinforcement Learning Outperforms Supervised Fine-Tuning: A Case Study on Audio Question Answering},
  author={Li, Gang and Liu, Jizhong and Dinkel, Heinrich and Niu, Yadong and Zhang, Junbo and Luan, Jian},
  journal={arXiv preprint arXiv:2503.11197},
  year={2025}
}

@article{wu2025stepaudio2,
  title={{Step-Audio 2} Technical Report},
  author={Wu, Boyong and Yan, Chao and Hu, Chen and Yi, Cheng and Feng, Chengli and Tian, Fei and others},
  journal={arXiv preprint arXiv:2507.16632},
  year={2025}
}

@article{liu2025voxtral,
  title={Voxtral},
  author={Liu, Alexander H. and Ehrenberg, Andy and Lo, Andy and Denoix, Cl{\'e}ment and Barreau, Corentin and Lample, Guillaume and others},
  journal={arXiv preprint arXiv:2507.13264},
  year={2025}
}

@article{he2024meralion,
  title={{MERaLiON-AudioLLM}: Bridging Audio and Language with Large Language Models},
  author={He, Yingxu and Liu, Zhuohan and Sun, Shuo and Wang, Bin and Zhang, Wenyu and Zou, Xunlong and others},
  journal={arXiv preprint arXiv:2412.09818},
  year={2024}
}

@article{gemmateam2026gemma4,
  title={{Gemma 4} Technical Report},
  author={{Gemma Team}},
  journal={arXiv preprint arXiv:2607.02770},
  year={2026}
}

@article{ghosh2026afnext,
  title={{Audio Flamingo Next}: Next-Generation Open Audio-Language Models for Speech, Sound, and Music},
  author={Ghosh, Sreyan and Goel, Arushi and Jayakumar, Kaousheik and Koroshinadze, Lasha and Anand, Nishit and Kong, Zhifeng and others},
  journal={arXiv preprint arXiv:2604.10905},
  year={2026}
}

@article{cui2026minicpmo,
  title={{MiniCPM-o 4.5}: Towards Real-Time Full-Duplex Omni-Modal Interaction},
  author={Cui, Junbo and Xu, Bokai and Wang, Chongyi and Yu, Tianyu and Sun, Weiyue and Xu, Yingjing and others},
  journal={arXiv preprint arXiv:2604.27393},
  year={2026}
}

@inproceedings{he2026measuring,
  title={Measuring Audio's Impact on Correctness: Audio-Contribution-Aware Post-Training of Large Audio Language Models},
  author={He, Haolin and Du, Xingjian and Sun, Renhe and Dai, Zheqi and Xiao, Yujia and Yang, Mingru and others},
  booktitle={International Conference on Learning Representations},
  year={2026},
  note={arXiv:2509.21060}
}

@inproceedings{hsu2025audioaware,
  title={Reducing Object Hallucination in Large Audio-Language Models via Audio-Aware Decoding},
  author={Hsu, Tzu-wen and Lu, Ke-Han and Chiang, Cheng-Han and Lee, Hung-yi},
  booktitle={IEEE Automatic Speech Recognition and Understanding Workshop},
  year={2025},
  note={arXiv:2506.07233}
}

@inproceedings{lin2026contrastive,
  title={How Contrastive Decoding Enhances Large Audio Language Models},
  author={Lin, Tzu-Quan and Huang, Wei-Ping and Lin, Yi-Cheng and Lee, Hung-yi},
  booktitle={IEEE Spoken Language Technology Workshop},
  year={2026},
  note={arXiv:2603.09232}
}

@inproceedings{li2016learning,
  title={Learning without Forgetting},
  author={Li, Zhizhong and Hoiem, Derek},
  booktitle={European Conference on Computer Vision},
  year={2016},
  note={arXiv:1606.09282}
}

@inproceedings{held2024diva,
  title={Distilling an End-to-End Voice Assistant Without Instruction Training Data},
  author={Held, William and Zhang, Yanzhe and Li, Minzhi and Shi, Weiyan and Ryan, Michael J. and Yang, Diyi},
  booktitle={Annual Meeting of the Association for Computational Linguistics},
  year={2025},
  note={arXiv:2410.02678}
}

@article{lu2025desta,
  title={{DeSTA2.5-Audio}: Toward General-Purpose Large Audio Language Model with Self-Generated Cross-Modal Alignment},
  author={Lu, Ke-Han and Chen, Zhehuai and Fu, Szu-Wei and Yang, Chao-Han Huck and Huang, Sung-Feng and Yang, Chih-Kai and others},
  journal={IEEE Transactions on Audio, Speech and Language Processing},
  volume={34},
  pages={2062--2076},
  year={2026},
  note={arXiv:2507.02768}
}

@inproceedings{lu2025speechifeval,
  title={{Speech-IFEval}: Evaluating Instruction-Following and Quantifying Catastrophic Forgetting in Speech-Aware Language Models},
  author={Lu, Ke-Han and Kuan, Chun-Yi and Lee, Hung-yi},
  booktitle={Interspeech},
  year={2025},
  note={arXiv:2505.19037}
}

@inproceedings{hu2022lora,
  title={{LoRA}: Low-Rank Adaptation of Large Language Models},
  author={Hu, Edward J. and Shen, Yelong and Wallis, Phillip and Allen-Zhu, Zeyuan and Li, Yuanzhi and Wang, Shean and Wang, Lu and Chen, Weizhu},
  booktitle={International Conference on Learning Representations},
  year={2022},
  note={arXiv:2106.09685}
}

@inproceedings{mauch2014pyin,
  title={{pYIN}: A Fundamental Frequency Estimator Using Probabilistic Threshold Distributions},
  author={Mauch, Matthias and Dixon, Simon},
  booktitle={IEEE International Conference on Acoustics, Speech and Signal Processing},
  pages={659--663},
  year={2014}
}

@inproceedings{panayotov2015librispeech,
  title={{LibriSpeech}: An {ASR} Corpus Based on Public Domain Audio Books},
  author={Panayotov, Vassil and Chen, Guoguo and Povey, Daniel and Khudanpur, Sanjeev},
  booktitle={IEEE International Conference on Acoustics, Speech and Signal Processing},
  pages={5206--5210},
  year={2015}
}

@inproceedings{nguyen2023expresso,
  title={{EXPRESSO}: A Benchmark and Analysis of Discrete Expressive Speech Resynthesis},
  author={Nguyen, Tu Anh and Hsu, Wei-Ning and D'Avirro, Antony and Shi, Bowen and Gat, Itai and Fazel-Zarani, Maryam and Remez, Tal and Copet, Jade and Synnaeve, Gabriel and Hassid, Michael and Kreuk, Felix and Adi, Yossi and Dupoux, Emmanuel},
  booktitle={Interspeech},
  pages={4823--4827},
  year={2023}
}

@misc{weinberger2015accent,
  title={{Speech Accent Archive}},
  author={Weinberger, Steven},
  year={2015},
  howpublished={George Mason University. \url{http://accent.gmu.edu}}
}

@inproceedings{gournay2018cafe,
  title={A {Canadian French} Emotional Speech Dataset},
  author={Gournay, Philippe and Lahaie, Olivier and Lefebvre, Roch},
  booktitle={ACM Multimedia Systems Conference},
  pages={399--402},
  year={2018}
}

@misc{pichorafuller2020tess,
  title={{Toronto} Emotional Speech Set ({TESS})},
  author={Pichora-Fuller, M. Kathleen and Dupuis, Kate},
  year={2020},
  howpublished={Borealis dataset. \url{https://doi.org/10.5683/SP2/E8H2MF}}
}

@article{adigwe2018emovdb,
  title={The Emotional Voices Database: Towards Controlling the Emotion Dimension in Voice Generation Systems},
  author={Adigwe, Adaeze and Tits, No{\'e} and El Haddad, Kevin and Ostadabbas, Sarah and Dutoit, Thierry},
  journal={arXiv preprint arXiv:1806.09514},
  year={2018}
}

@inproceedings{james2018jlcorpus,
  title={An Open Source Emotional Speech Corpus for Human Robot Interaction Applications},
  author={James, Jesin and Tian, Li and Watson, Catherine Inez},
  booktitle={Interspeech},
  pages={2768--2772},
  year={2018}
}

@inproceedings{richter2024ears,
  title={{EARS}: An Anechoic Fullband Speech Dataset Benchmarked for Speech Enhancement and Dereverberation},
  author={Richter, Julius and Wu, Yi-Chiao and Krenn, Steven and Welker, Simon and Lay, Bunlong and Watanabe, Shinji and Richard, Alexander and Gerkmann, Timo},
  booktitle={Interspeech},
  pages={4873--4877},
  year={2024}
}

@article{fonseca2022fsd50k,
  title={{FSD50K}: An Open Dataset of Human-Labeled Sound Events},
  author={Fonseca, Eduardo and Favory, Xavier and Pons, Jordi and Font, Frederic and Serra, Xavier},
  journal={IEEE/ACM Transactions on Audio, Speech, and Language Processing},
  volume={30},
  pages={829--852},
  year={2022}
}

@article{rashid2023nonspeech7k,
  title={{Nonspeech7k} Dataset: Classification and Analysis of Human Non-Speech Sound},
  author={Rashid, Muhammad Mamunur and Li, Guiqing and Du, Chengrui},
  journal={IET Signal Processing},
  volume={17},
  number={6},
  pages={e12233},
  year={2023}
}

@inproceedings{piczak2015esc,
  title={{ESC}: Dataset for Environmental Sound Classification},
  author={Piczak, Karol J.},
  booktitle={ACM International Conference on Multimedia},
  pages={1015--1018},
  year={2015}
}

@inproceedings{zhang2021speechocean,
  title={speechocean762: An Open-Source Non-Native {English} Speech Corpus for Pronunciation Assessment},
  author={Zhang, Junbo and Zhang, Zhiwen and Wang, Yongqing and Yan, Zhiyong and Song, Qiong and Huang, Yukai and Li, Ke and Povey, Daniel and Wang, Yujun},
  booktitle={Interspeech},
  pages={3710--3714},
  year={2021}
}

@article{shankar2024jibo,
  title={The {JIBO} Kids Corpus: A Speech Dataset of Child-Robot Interactions in a Classroom Environment},
  author={Shankar, Natarajan Balaji and Afshan, Amber and Johnson, Alexander and Mahapatra, Aurosweta and Martin, Alejandra and Ni, Haolun and Park, Hae Won and Perez, Marlen Quintero and Yeung, Gary and Bailey, Alison and Breazeal, Cynthia and Alwan, Abeer},
  journal={JASA Express Letters},
  volume={4},
  number={11},
  pages={115201},
  year={2024},
  doi={10.1121/10.0034195}
}

@misc{hopkins2019aru,
  title={{ARU} Speech Corpus ({University of Liverpool})},
  author={Hopkins, Carl and Graetzer, Simone and Seiffert, Gary},
  year={2019},
  howpublished={University of Liverpool Data Catalogue},
  doi={10.17638/datacat.liverpool.ac.uk/681}
}

@misc{kennedy2016children,
  title={Children Speech Recording ({English}, Spontaneous Speech + Pre-Defined Sentences)},
  author={Kennedy, James and Lemaignan, S{\'e}verin and Montassier, Caroline and Lavalade, Pauline and Irfan, Bahar and Papadopoulos, Fotios and Senft, Emmanuel and Belpaeme, Tony},
  year={2016},
  howpublished={Zenodo},
  doi={10.5281/zenodo.200495}
}

@article{olatunji2023afrispeech,
  title={{AfriSpeech-200}: Pan-{African} Accented Speech Dataset for Clinical and General Domain {ASR}},
  author={Olatunji, Tobi and Afonja, Tejumade and Yadavalli, Aditya and Emezue, Chris Chinenye and Singh, Sahib and Dossou, Bonaventure F. P. and Osuchukwu, Joanne and Osei, Salomey and Tonja, Atnafu Lambebo and Etori, Naome and Mbataku, Clinton},
  journal={Transactions of the Association for Computational Linguistics},
  volume={11},
  pages={1669--1685},
  year={2023}
}

@inproceedings{kalluri2021nisp,
  title={{NISP}: A Multi-lingual Multi-accent Dataset for Speaker Profiling},
  author={Kalluri, Shareef Babu and Vijayasenan, Deepu and Ganapathy, Sriram and M, Ragesh Rajan and Krishnan, Prashant},
  booktitle={IEEE International Conference on Acoustics, Speech and Signal Processing},
  pages={6953--6957},
  year={2021}
}

@inproceedings{scheibler2018pyroomacoustics,
  title={Pyroomacoustics: A {Python} Package for Audio Room Simulation and Array Processing Algorithms},
  author={Scheibler, Robin and Bezzam, Eric and Dokmani{\'c}, Ivan},
  booktitle={IEEE International Conference on Acoustics, Speech and Signal Processing},
  pages={351--355},
  year={2018}
}

@inproceedings{mcauliffe2017mfa,
  title={{Montreal Forced Aligner}: Trainable Text-Speech Alignment Using {Kaldi}},
  author={McAuliffe, Michael and Socolof, Michaela and Mihuc, Sarah and Wagner, Michael and Sonderegger, Morgan},
  booktitle={Interspeech},
  pages={498--502},
  year={2017}
}

@article{zhao2023librisqa,
  title={{LibriSQA}: A Novel Dataset and Framework for Spoken Question Answering with Large Language Models},
  author={Zhao, Zihan and Jiang, Yiyang and Liu, Heyang and Wang, Yu and Wang, Yanfeng},
  journal={IEEE Transactions on Artificial Intelligence},
  volume={6},
  number={11},
  pages={2884--2895},
  year={2025},
  note={arXiv:2308.10390}
}

@inproceedings{rafailov2023direct,
  title={Direct Preference Optimization: Your Language Model Is Secretly a Reward Model},
  author={Rafailov, Rafael and Sharma, Archit and Mitchell, Eric and Manning, Christopher D. and Ermon, Stefano and Finn, Chelsea},
  booktitle={Advances in Neural Information Processing Systems},
  year={2023},
  note={arXiv:2305.18290}
}

@inproceedings{cuervo2026closing,
  title={Closing the Gap Between Text and Speech Understanding in {LLMs}},
  author={Cuervo, Santiago and Seto, Skyler and de Seyssel, Maureen and Bai, Richard He and Gu, Zijin and Likhomanenko, Tatiana and Jaitly, Navdeep and Aldeneh, Zakaria},
  booktitle={International Conference on Learning Representations},
  year={2026},
  note={arXiv:2510.13632}
}

@inproceedings{fan2026melllm,
  title={{LLM} Can Read Spectrogram: Encoder-free Speech-Language Modeling},
  author={Fan, Ruchao and Wang, Yiming and Hu, Yuxuan and Ren, Bo and Xia, Yufei and Wang, Xiaofei and Qian, Yao and Liu, Shujie and Li, Jinyu},
  booktitle={IEEE Spoken Language Technology Workshop},
  year={2026},
  note={arXiv:2606.10231}
}

@article{peng2026represented,
  title={Represented but Ignored: A Causal Account of Prosodic Underuse in Audio-Language Models},
  author={Peng, Linkai and Nuchged, Baorian},
  journal={arXiv preprint arXiv:2608.19211},
  year={2026}
}

@misc{lmmslab2025aero,
  title={{Aero-1-Audio} model card},
  author={{LMMs-Lab}},
  year={2025},
  howpublished={Hugging Face. \url{https://huggingface.co/lmms-lab/Aero-1-Audio}}
}
